\documentclass{IEEEtaes}

\usepackage{color,array,amsthm}
\usepackage{graphicx}
\usepackage{amsmath} 
\usepackage{subfigure}
\usepackage{algorithm} 
\usepackage{algpseudocode}
\usepackage{makecell}

\usepackage{booktabs}

\usepackage{caption}
\usepackage{float} 

\usepackage{pifont}

\jvol{XX}
\jnum{XX}
\jmonth{XXXXX}
\paper{1234567}
\pubyear{2022}
\doiinfo{TAES.2022.Doi Number}

\begin{document}

\title{Dynamic real-time multi-UAV cooperative mission planning method under multiple constraints
}

\author{Chenglou Liu}

\author{Yufeng Lu}

\author{Fangfang Xie}

\author{Tingwei Ji}

\author{Yao Zheng}
\affil{Zhejiang University, Hangzhou, china} 




\authoraddress{Chenglou Liu, Yufeng Lu, Fangfang Xie, Tingwei Ji and Yao Zheng are with School of Aeronautics and Astronautics, Zhejiang University, Hangzhou 310027, China, 
(e-mail: {12424006@zju.edu.cn}, {22024035@zju.edu.cn}, {fangfang\_xie@zju.edu.cn}, {zjjtw@zju.edu.cn}, {yao.zheng@zju.edu.cn}). (corresponding author: Tingwei Ji.)}


\maketitle

\begin{abstract}As UAV popularity soars, so does the mission planning associated with it. 
The classical approaches suffer from the triple problems of decoupled of task assignment and path planning, poor real-time performance and limited adaptability. Aiming at these challenges, this paper proposes a dynamic real-time multi-UAV collaborative mission planning algorithm based on Dubins paths under a distributed formation structure. 
Dubins path with multiple advantages bridges the gap between task assignment and path planning, leading to a coupled solution for mission planning. 
Then, a series of acceleration techniques, task clustering preprocessing, highly efficient distance cost functions, low-complexity and less iterative task allocation strategies, are employed to guarantee the real-time performance of the algorithms. 
To cope with different emergencies and their simultaneous extremes, real-time planning of emerging tasks and mission replanning due to the reduction of available UAVs are appropriately handled.
Finally, the developed algorithm is comprehensively exemplified and studied through simulations, 
highlighting that the proposed method only sacrifices 9.57\% of the path length, while achieving a speed improvement of 4-5 orders of magnitude over the simulated annealing method, with a single mission planning of about 0.0003s.

\end{abstract}


\section{INTRODUCTION}
Nowadays, the application of Unmanned Aerial Vehicles (UAVs) in military and civilian is exponentially increasing [1]-[3]. Specifically, fixed-wing UAVs have attracted the attention of scholars due to their advantages of long endurance and heavy payload [4]-[6]. However, as the environment and task requirements become increasingly sophisticated, a single fixed-wing UAV can no longer meet the practical demands, leading to the emergence of UAV swarms to compensate for its shortcomings. The collaboration of multiple UAVs is essential for accomplishing complex and dynamic missions, with effective mission planning being a prerequisite for successful multi-UAV collaboration [7]. Multi-UAV collaborative mission planning refers to obtaining reasonable task allocation and path planning based on the observed environmental situation and target status, combined with the UAV's own attributes and status. It also includes the ability to negotiate and adjust the mission plan in real time as the environment changes dynamically. Therefore, multi-UAV collaborative mission planning comprises two components: collaborative task allocation and coordinated path planning for multiple UAVs. Although having been studied, multi-UAV collaborative mission planning still remains some attractive challenges to address.

First, existing researches usually decouple mission planning into two distinct components: task allocation and path planning. 
Task allocation precedes path planning, which finalizes the assignment of tasks and traversal sequence for each UAV.
The outcomes of task allocation serve as the input for path planning. 
Path planning specifies the entry and exit angles of targets, and smooths the path to satisfy non-holonomic motion constraints.
The primary reason for the decoupling of mission planning in classical methods is that the applied distance cost is the Euclidean distance.
However, straight paths are clearly not flyable paths under target heading angles and kinematic constraints. This makes path planning independent of and after task allocation.
In fact, the allocation of tasks and the sequence of traversal directly affect the planned path. 
In turn, the length of the planned path can also influence task allocation.
Two processes mutually impact each other. 
Although the decoupling process simplifies the problem, it also induces issues such as the inability to account for heading angles, low efficiency, and non-optimal solutions.


Secondly, real-time mission planning is a fundamental cornerstone not only for autonomous flight but also for subsequent re-planning in emergency situations.
The goal of real-time mission planning is to plan the appropriate mission to each UAV while keeping the planning time as short as possible within an acceptable total distance.
On one hand, the distinctive feature is the absence of a sequential list of tasks and corresponding paths before the UAV departs. 
This means that the UAV's next task and path are instantly planned based on the current position and attitude after the UAV completes the current task.
On the other hand, real-time mission replanning is crucial for enhancing mission success when sudden changes occur in the target or external environment during flight. 
Compared with previous methods, real-time mission planning must address the challenges posed by dynamic environment and uncertainty to deliver effective and feasible scheduling results within limited time and computational resources, thereby increasing the complexity and difficulty of the solution.

Finally, the UAV's ability to process emergencies can significantly improve the success rate of the mission, enhance environmental adaptability and ensure the robustness of UAVs.
However, the scenarios encountered by UAVs in existing studies are mostly static, while assuming that the UAVs perform their tasks flawlessly and with excellence. 
This is not compatible with the real world, where the real environment is dynamic and variable.
On one hand, as the environment changes, unexpected new tasks frequently appear. 
This necessitates that UAV swarm quickly senses new tasks, incorporates them into the task list, and assigns UAVs to complete these new tasks. 
On the other hand, various emergencies, such as communication failures and mechanical failures, may occur during flight.
These adverse situations result in damage to the UAV, preventing it from continuing to fulfill its assigned tasks. 
Therefore, the remaining tasks should be carried out with the assistance of other UAVs.
Moreover, even the few studies that address dynamic problems typically consider different contingencies separately, ignoring extreme cases where multiple contingencies occur simultaneously.

\begin{figure}[h]
\centering
\includegraphics[width=0.9\linewidth]{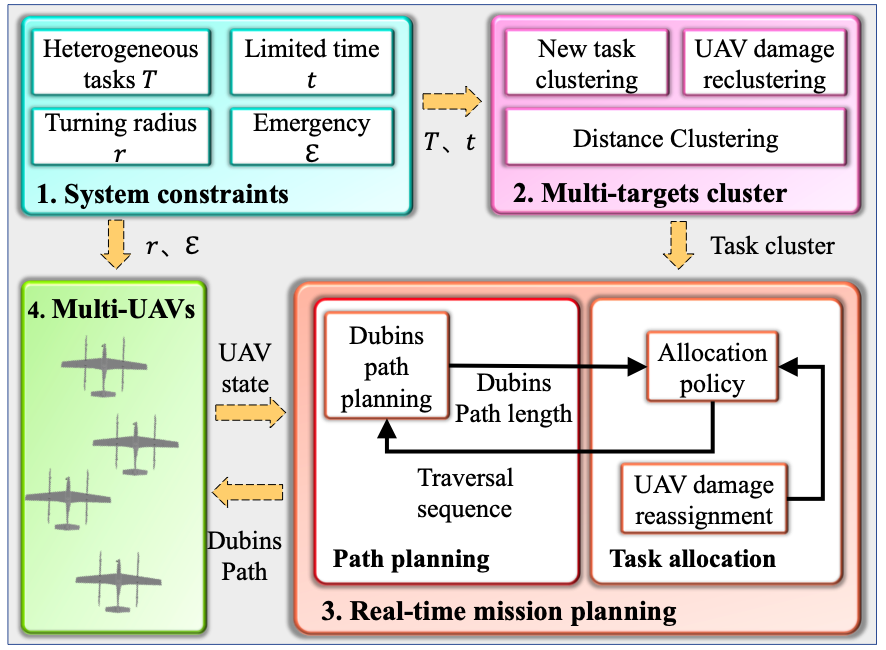}
\caption{Overall structure flowchart of the proposed algorithms.}
\label{myfigure29}
\end{figure}

Aiming at the above issues, this paper explores the multi-UAV real-time collaborative mission planning and the response strategies for emergencies,as shown in Fig. \ref{myfigure29}.
Real-time collaborative mission planning is to find the assignment of UAVs to tasks and the corresponding flight paths such that (1) task assignment and path planning are addressed coupled, and (2) the planning time is as short as possible while ensuring an acceptable total distance.
The goal of planning for emergencies is to determine reasonable allocation such that (1) the single allocation time is as short as possible, and (2) various emergencies are handled appropriately. 
The contributions of this paper are summarized below.

1) Distinguishing from previous decoupling methods, this paper proposes a coupled task assignment and path planning solution.
Specifically, the distance cost in the task assignment phase is the Dubins path distance derived from path planning, while the final adopted path is determined by the result of task assignment. 
What is crucial here is that the Dubins paths employed for decision and flight are identical.
This is equivalent to combining the two processes of task allocation and path planning into one, significantly reducing planning time.
The adopted Dubins path allows the proposed algorithm not only to take into account the task point heading angle constraints as early as the task assignment phase but also to be applicable to scenarios involving heterogeneous tasks.
Furthermore, this paper demonstrates through numerical experiments that employing straight-line distance and Dubins path distance yields different mission planning schemes. 

2) To achieve real-time mission planning, this paper implements the following measures to accelerate the process. 
First, the speed of calculating the distance cost directly influences the efficiency of task assignment. 
By reducing the types of Dubins path solutions and simplifying the solution format, Dubins paths are generated in an order of magnitude $10^{-4}s$. Then, computationally lightweight task allocation strategies, such as greedy and Hungarian algorithms, are selected.
The tedious iteration process is also rejected, retaining only low-level, computationally light iterations. 
Finally, the size of the decision space also directly influences the time for UAVs to make decisions. 
The pre-processing of clustering is employed to effectively partition the original large decision space into smaller subspaces to be allocated to each UAV, thereby accelerating the decision-making process.

3) To enhance the adaptability of the UAV, we implement coping strategies aligned with the proposed algorithms to address various emergencies, including new tasks and UAV failures. Furthermore, we consider the extreme scenario where new tasks and UAV failures occur simultaneously and propose appropriate coping strategies.

The paper is organized as follows.
Section II introduces the works related to mission planning.
Section III not only describes and models the problem, but also introduces the heterogeneous tasks and the distance cost function.
The characterization and procedure of the proposed method are described in detail in Section IV. 
In Section V, response strategies for new tasks, UAV damage, and their combination scenarios are discussed.
Section VI evaluates the effectiveness and performance of the proposed algorithm through simulation experiments. 
Section VII summarizes the conclusions.

\section{RELATED WORK}
Multi-UAV collaborative mission planning aims to control a group of UAVs to reach the target along a suitable path and complete the mission through the coordination and cooperation of UAVs. This process is a typical multi-constraint combinatorial optimization problem [8]. 

To study the problem rigorously, it is essential to mathematically abstract the actual issue and accurately represent it with an appropriate model. 
Common mathematical models for mission planning include the multiple traveling salesman problem (MTSP) [9-10], the vehicle routing problem (VRP) [11-12], and the mixed-integer linear programming problem (MILP) [13-14].
MTSP is the classical mathematical model describing task assignment for UAVs. In recent years, to adapt it for fixed-wing UAVs, the assumption that the vehicle’s motion constraints adhere to the Dubins model has been incorporated into the path planning solution. 
This led to a new variant known as MDTSP [15]. 
VRP is a constrained extension of the well-known Traveling Salesman Problem (TSP), enabling more detailed modeling of real-world scenarios. 
Similarly, MILP is a popular modeling methodology that converts the multi-UAV collaborative mission planning into an integer programming problem, allowing for the introduction of various constraints. 

Once the model is established, it can be solved adopting appropriate optimization methods. 
To our knowledge, numerous methods with different principles have been proposed. It is beyond the scope of this paper to cover all the existing studies, so we focus on methods that are either representative or directly related to the problem addressed herein. 
According to the UAV formation control mode, the algorithms can be classified into centralized and distributed. Additionally, depending on whether the solution is optimal, the algorithms can be categorized as deterministic or heuristic. To clearly present the relevant methods, this paper adopts the second categorization while also illustrating their control modes.

Optimization algorithms, specifically deterministic algorithms, aim to provide optimal solutions to mission planning problems. Deterministic algorithms include branching algorithms and their variants, dynamic programming algorithms, Hungarian algorithms, and others. Shima et al. [16] formulated the collaborative task allocation problem as a decision tree and subsequently solved it using a branch-and-bound optimization algorithm to obtain an optimal solution. 
Alighanbari et al. [17] investigated the assignment problem for UAVs in the presence of environmental risks, formulated it as a dynamic programming problem, and proposed two approximations: one-step and two-step forward-looking. 
The Hungarian algorithm is another classical algorithm which treats the task allocation problem as a combinatorial optimization problem using graph theory and solves the problem in polynomial time. 
The algorithm computes an estimate of each agent's utility to maximize overall utility. 
Samiei et al. [18] proposed a Hungarian-based  method to address the task allocation problem where the number of tasks exceeds the number of agents, comparing its performance with that of a consensus-based bundling algorithm. 
This work did not take into account the kinematic constraints of the UAV, which makes the path unflyable for fixed-wing UAV.

Heuristic algorithms aim to deliver efficient and feasible suboptimal solutions for task allocation. Heuristic methods include simulated annealing (SA), genetic algorithms, ant colony optimization, particle swarm optimization, greedy algorithms, auction algorithms, and others.
When faced with a large number of locally optimal solutions, simulated annealing is highly effective at locating globally optimal solutions. 
Wang et al. [19] modeled a vehicle routing problem with specific time windows applying mixed-integer programming model and developed a parallel simulated annealing method to solve it. 
Wang et al. [20] suggested a multi-UAV reconnaissance tasking model that considers heterogeneous tasks and UAV constraints, along with an improved genetic algorithm used to solve the MDTSP problem. However, this research did not account for sudden emergencies.
Addressing the challenges of finite duration and spatial instability of time-sensitive targets, Zhang et al. [21] proposed a dynamic multi-UAV mission planning algorithm based on a time window mechanism and solved it using an improved particle swarm algorithm. 
Auction algorithm (AA) is a distributed algorithm for solving allocation problems. Choi et al. [22] pioneered the Consensus-Based Auction Algorithm (CBAA). They extended their algorithm to address the multi-allocation problem of assigning multiple task sequences to each agent by proposing the Consensus-Based Bundling Algorithm (CBBA). 
Wu et al. [23] proposed an extension of CBBA that deals with the coupling of task allocation and path planning optimization in scenarios with multi-task demands and time constraints. Moreover, they investigated a new distributed genetic algorithm to enhance communication protocols for formation.

In addition, there are a variety of excellent works that discuss emergencies and parameter uncertainty, such as new targets and UAV damage. 
To ensure load balancing of the UAV, reduce overall task completion time, and lower the energy consumption of the entire system, Wu et al. [24] proposed an improved self-organizing mapping (ISOM) algorithm for efficient multi-task allocation. Based on this, they integrated the attention mechanism with the ISOM algorithm to cope with emergencies. 
Gao et al. [25] focused on the task reassignment problem in emergency situations and proposed a task reassignment algorithm based on contractual network protocols. 
Chen et al. [26] introduced robust optimization methods, dyadic theory, and a novel combinatorial algorithm to address the parameter uncertainty problem. Additionally, to solve online problems involving time-sensitive uncertainty, they proposed a practical online hierarchical planning algorithm.

In summary, on the one hand, existing research on UAV mission planning divides task assignment and path planning into two separate components. 
Some studies only focus on task allocation and ignore path planning, which makes the path unflyable. 
The extensive reliance on heuristic algorithms significantly hampers their real-time performance. 
On the other hand, much of the literature does not consider the management of unexpected events, resulting in a less robust system with limited autonomy and flexibility. Once a single UAV is damaged, it will lead to mission failure. The limited studies addressing dynamic problems also overlook extreme cases in which multiple contingencies occur simultaneously. 
Although each of the complex aspects mentioned above has been partially addressed in the literature, there has been limited work that systematically combines them all. Therefore, this paper explores a real-time collaborative task assignment method based on Dubins path distance. 

\section{PRELIMINARIES AND PROBLEM FORMULATION}
\label{PRELIMINARIES AND PROBLEM FORMULATION}
Under the premise of ensuring the validity of the model, reasonable assumptions are made to reduce complexity. 

\textbf{Assumption 1:} 
The fixed-wing UAV is modeled by a planar nonholonomic vehicle, generally known as a unicycle. 
The UAVs have collision-free paths by flying at a given altitude layering.
The UAVs are flying at a constant speed during the flight.
UAVs have sufficient fuel to meet range constraints.

\textbf{Assumption 2:} In the mission scenarios discussed in this paper, obstacles and hazardous areas are currently not taken into account. 

\subsection{Problem Statement}
Consider multiple UAVs forming a collaborative swarm flying in a 3D aerial environment and performing different operations against ground targets. Suppose a set of static targets $\mathcal{T}=\left\{T_{1}, T_{2},\ldots, T_{N}\right\}$ with known locations is randomly distributed on the ground within the mission area. While a group of homogeneous UAV $\mathcal{U} = \{U_1, U_2, \ldots, U_K\}$ is centrally docked at the base station $T_{0}$, awaiting their tasks, where the total number of $\mathcal{T}$ and $\mathcal{U}$ are $N$ and $K$, respectively. It is also assumed that $K < N$, which is commonly encountered in many practical applications. 
The purpose of mission planning is to assign tasks and plan paths in real time without a predefined list of tasks and corresponding paths, so that each task is visited by a UAV once and only once and the total distance is as small as possible.
Finally, the mission sequence $\mathcal{D}_{k}$ formed by the UAV should satisfy the following equation:
\begin{equation}
\label{myequation1}
\mathcal{D}_{k}=\left\{ T_{i} \mid T_{i}\in\mathcal{T}, x_{i, j}^{k}=1\right\},
\end{equation}
\begin{equation}
\label{myequation2}
\mathcal{T}=\bigcup_{k=1}^{K}\mathcal{D}_{k},
\end{equation}
\begin{equation}
\label{myequation3}
\mathcal{D}_{k}\bigcap\mathcal{D}_{l}=\emptyset, \quad \forall k, l \text{ and } k \neq l.
\end{equation}
Where $x_ {i,j}^k$ is a decision variable that indicates whether the UAV $U_k$ travels from the target $T_i$ to the target $T_j$. Furthermore, $N_k$ and $L_k$ denote the number of targets and the length of the trip in sequence $\mathcal{D}_{k}$, respectively.
\begin{equation}
\label{myequation6}
x_{i,j}^k = 
\begin{cases} 
1, & \text{if } U_k \text{ executes } T_{i,j}, \\
0, & \text{otherwise}.
\end{cases}
\end{equation}

\textbf{Vehicles.} The attributes of a UAV include configuration and status, represented by $U_{att} = (\mathbf{p}, U_{state})$.
$\mathbf{p} = (x, y, \theta)$  denotes the configuration of the UAV, including its horizontal position in Cartesian inertial coordinates $(x,y)$ and the heading angle $\theta$, which increases in the counterclockwise direction. 
The Dubins model is introduced to establish the kinematics of the UAV and to generate the fundamental flight path.
Additionally, $U_{state}$ represents the state of the UAV, and the four attributes of the UAV can be described by the following equation:
\begin{equation}
\label{myequation4}
U_{\text{state}} = 
\begin{cases}
0, & \text{idle}, \\
1, & \text{in-transit}, \\
2, & \text{busy}, \\
3, & \text{damaged}.
\end{cases}
\end{equation}
The idle status refers that the UAV is not currently receiving tasks and is in a standby mode.
The in-transit status indicates to the UAV flying along a connecting path.
In other words, it is traveling from the previous task point to the next.
During this phase, the UAV neither engages in task allocation nor receives any path information. 
The busy status signifies that the UAV has arrived at the mission point, and depending on the type of mission, the UAV needs to perform various tasks.
As a result, there is great variation in the time UAVs spend on missions.
Additionally, UAV may enter a special status known as damaged. As the name implies, this state typically indicates that the UAV has been physically damaged or has lost communication due to unforeseen circumstances, rendering it unable to fly.

\textbf{Mission.} 
Similarly, to better describe the properties of the task, it is represented by the $T_{att} = (\mathbf{x}, T_{\text{state}}, T_{\text{type}})$. $\mathbf{x}=(x,y)$ denotes the horizontal position of the task on the ground. $T_{\text{type}}$ represents the state of the task, which is illustrated in the following equation:
\begin{equation}
\label{myequation5}
T_{state} = 
\begin{cases}
0, & \text{unassigned}, \\
1, & \text{assigned}, \\
2, & \text{completed}.
\end{cases}
\end{equation}
The unassigned status refers that the task is not currently assigned to any UAV and has yet to be completed. Tasks in this state constitute the decision space for the UAVs and will be allocated to the appropriate UAV at the right time according to a specific policy. 
The assigned state indicates that the task has been assigned to a specific UAV, which may be either in-transit or actively performing the task. A task in the assigned state does not appear in the UAV's decision space. 
The completed status means that the particular UAV has reached mission point and has planned the appropriate coverage paths according to the requirements of the task, thus completing the task in its entirety. Completed tasks also do not appear in the UAV's decision space.

$T_{\text{type}}$ denotes the type of task, and different task types have varying connection and coverage paths. See below for a detailed description.

\subsection{Integer Programming Formula}
The real-time collaborative mission planning problem addressed in this paper can be modeled as MTSP.
Each feasible solution must satisfy a set of mission, vehicle, and environment constraints. 
The MTSP is transformed into an equivalent assignment-based double-exponential integer linear programming problem using the binary decision variables $x_{i,j}^{k} \in \{0,1\}$ introduced above. 
In double-exponential integer linear programming, the objective function is the total path length of each UAV's movement. 
The path of that UAV comprises the following four parts: (1) the path from the base to the entry configuration of the first target; (2) all connected paths from the exit configuration of the current target $T_j$ to the entry configuration of the next target $T_{j+1}$; 
(3) the coverage paths of targets, its definition will be provided later; 
and (4) the path from the exit configuration of the last target to the base station. Therefore, the total length of the UAV path is calculated as follows:
\begin{equation}
\label{myequation7}
L^{(1)}_{k} = L_{0,j}^k([x_0, \theta_0], [x_j, \theta_j])x_{0,j}^{k}, j\in(1,..,N)
\end{equation}
\begin{equation}
\label{myequation8}
L^{(2)}_k = \sum_{i=1}^{N-1} \sum_{j=1}^{N-1} L_{i,j}^k \left( [x_i', \theta_i'], [x_{j}, \theta_{j}] \right)x_{i,j}^{k},
\end{equation}
\begin{equation}
\label{myequation9}
L^{(3)}_k = \sum_{i=1}^{N-1} \sum_{j=1}^{N-1} l_j x_{i,j}^{k},
\end{equation}
\begin{equation}
\label{myequation10}
L^{(4)}_k = L_{j,0}^k ([x_j', \theta_j'], [x_0, \theta_0]) x_{j,0}^{k}, j\in(1,..,N)
\end{equation}
\begin{equation}
\label{myequation11}
L_k = L^{(1)}_k + L^{(2)}_k + L^{(3)}_k + L^{(4)}_k.
\end{equation}
Where $[\mathbf{x}_0, \theta_0]$ is the configuration of UAV at the base; $[\mathbf{x}_j, \theta_j]$ is the entry configuration $P_j$ of the target $T_j$; $[\mathbf{x}_i', \theta_i']$, $[\mathbf{x}_j', \theta_j']$ is the exit configuration of the target$T_i$, $T_j$; and $l_j$ is the length of the coverage path of the target $T_j$.

Thus, the double exponential integer linear programming problem is
\begin{equation}
\label{myequation12}
\min \sum_{k=1}^{K} L_k
\end{equation}
Subject to 
\begin{equation}
\label{myequation13}
\sum_{j=1}^{N} \sum_{k=1}^{K} x_{0,j}^{k} = K
\end{equation}
\begin{equation}
\label{myequation14}
\sum_{j=1}^{N} \sum_{k=1}^{K} x_{j,0}^{k} = K
\end{equation}
\begin{equation}
\label{myequation15}
\sum_{j=0}^{N} \sum_{k=1}^{K} x_{i,j}^{k} = 1, i = 2, \ldots, N,
\end{equation}
\begin{equation}
\label{myequation16}
\sum_{i=0}^{N} \sum_{k=1}^{K} x_{i,j}^{k} = 1, j = 2, ..., N,
\end{equation}


\begin{equation}
\label{myequation18}
\theta_i^{in}  = \phi_i, \phi_i \in [0, 2\pi] \text{ or } \phi_i = \emptyset, i = 1, 2, ..., N
\end{equation}
\begin{equation}
\label{myequation19}
\sum_{i=1}^{N} \sum_{j=1}^{N} x_{i,j}^k \geq 2
\end{equation}
Constraints (13) and (14) ensure that exactly $K$ UAVs leave and return to the starting point. Constraints (15) and (16) ensure that each mission point is visited once and only once. 
Constraint (17) indicates the presence of a heading angle constraint at the mission point. $\phi_i$ takes a null value to indicate that there is no heading angle constraint, while $\phi_i$ takes a radian value to indicate the presence of a heading angle constraint. 
Constraint (18) ensures that the total number of tasks assigned to the k-th UAV is not excessively small. 

\subsection{Heterogeneous Missions}
\begin{figure}[h]
\centering
\includegraphics[width=0.75\linewidth]{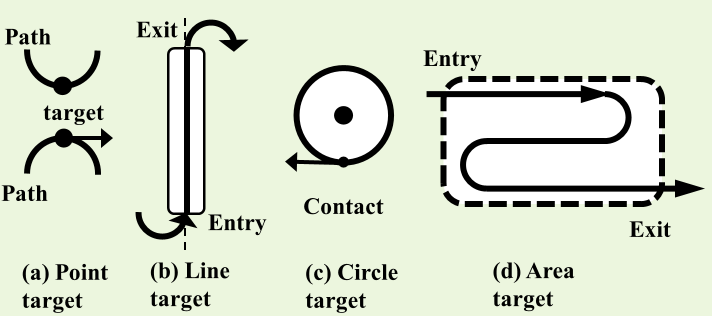}
\caption{Different types of task. A target with an arrow indicates that the target has a heading angle constraint}
\label{myfigure1}
\end{figure}

Much of the research on task allocation has focused on the characteristics of UAVs, while the characteristics of targets are often overlooked or regarded as homogeneous. 
However, in real environments, there are various types of targets with different characteristics and shapes, such as buildings, motorways and squares.
Therefore, in this paper, targets are categorized into four types: point target, line target, circle target and area target, based on their geometry characteristics, as shown in Fig.\ref{myfigure1}.
Point targets are further categorized into point target $T_{point}$ without direction constraints and point target $T_{point}'$ with direction constraints based on whether they have a preset approach direction.
Obviously, different types of targets have different mission requirements.

Transitions between different types of missions require flyable paths. The connecting path $P_{connt}$ refers to the feasible path for the UAV to move from the mission point $T_i$ to the mission point $T_j$, which may be classified as either a CS-type or CSC-type Dubins path. This classification depends on whether there is a heading angle constraint at the entry point of the target mission. Upon reaching the target, the UAV must perform heterogeneous tasks based on the characteristics of the target. 
The coverage path $P_{cover}$ refers to the path planned by the UAV at the mission point based on the actual needs of the heterogeneous mission.
Depending on the mission, the shape of the coverage path varies, such as overflying the target point, straight-line traverses, circling, and Z-coverage.
\begin{table}[htbp]
    \begin{center}
    \caption{Characterization of connection paths and coverage paths for heterogeneous tasks, where '1' denotes CS-type path and '2' denotes CSC-type path.}
    \label{mytable1}
    \begin{tabular}{lcc}
        \toprule
        Mission type & Connecting path & Coverage path\\
        \midrule
        Unconstrained point target & 1 & flyover\\
        Constrained point target & 2 & flyover\\
        Linear target & 2 & straight-line \\
        Circular target & 2 & circular\\
        Area target & 2 & zigzag path\\
        \bottomrule
    \end{tabular}
    \end{center}
\end{table}

\subsection{Dubins Path Distance Cost}
The coverage paths are predefined and fixed. However, the connecting paths between task points exhibit significant variations depending on the task pair configuration. 
Thus, the connecting path is the distance cost to focus on.
After completing the previous task, the UAV naturally has a direction and may not be directed toward the next possible target, thereby increasing the heading angle limit at the current position.
Depending on whether there is an entry constraint for the next target, two types of two-point path planning problems arise.
One involves attitude constraints at both the UAV's starting and termination points, with representative task types including constrained point tasks. 
This path planning problem is identified as the classical Markov-Dubins problem, and its optimal solution is a CSC-type Dubins path. Where C denotes a circular segment and S denotes a straight segment.
Our previous work [27] proposed an efficient path construction method based on analytic geometry, which is not reiterated here. 


\begin{figure}[h]
\centering
\includegraphics[width=1\linewidth]{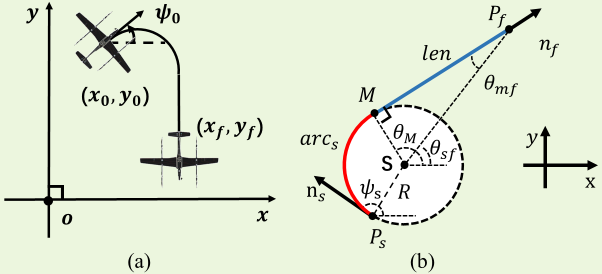}
\caption{Dubins path. (a) Geometric of the degenerate Markov-Dubins problem. (b) CS-type path construction geometric intuition. The red arc denotes the starting arc, the blue line denotes the intermediate tangent segment.}
\label{myfigure30}
\end{figure}


The other issue concerns a scenario in which only the starting point has attitude constraints, and the typical mission type is an unconstrained point target, as illustrated in Fig. \ref{myfigure30}(a).
On the one hand, the study [27] has demonstrated that the optimal path is a CS-type Dubins path with heading angle constraints at the starting point only; on the other hand, to further improve the computational efficiency, we adopt a more concise form of CS-type Dubins path.
Compared to the four CSC-type (LSL,LSR,RSL,RSR) paths, the CS-type paths have only two types, LS and RS, as shown in Fig. \ref{myfigure30}(b). 
Where R(L) stands for turning to the right (left) and S stands for straight flight. 
RS(LS) denotes the path is composed of arcs and lines to the right (left).

Considering the need for rapid construction of candidate optimal paths, this paper presents an analytic geometry-based method to generate CS-type Dubins paths. First, to obtain the path, it is essential to utilize the steering of the initial arc to determine the center of the starting circle $S$. This determines the position of the starting arc.
\begin{equation}
\label{myequation21}
(x_s, y_s) = \left( x_0 + R \cos \left( \theta_0 \pm \frac{\pi}{2} \right), y_0 + R \sin \left( \theta_0 \pm \frac{\pi}{2} \right) \right)
\end{equation}
Where $(x_0, y_0)$ is the starting point coordinate.
When the UAV turns left, the angle is taken as $\theta+\pi/2$, while the UAV turns right, the angle is taken as $\theta-\pi/2$. 

Then, the coordinate of the tangent point $M$ is derived. Based on the coordinates of the center of the starting circle and the endpoint, the angle $\theta_{sf}$ between the connecting line $SP_f$ and the X-axis can be calculated, as well as the length of the line segment $SP_f$.
\begin{equation}
\label{myequation22}
\theta_{sf} = \arctan \left( \frac{y_f - y_s}{x_f - x_s} \right)
\end{equation}
\begin{equation}
\label{myequation23}
\operatorname{len}_{SP_f} = \sqrt{(y_f - y_s)^2 + (x_f - x_s)^2}
\end{equation}
The angle between line segment $SP_f$ and tangent segment $MP_f$ is gained by the following equation,
\begin{equation}
\label{myequation24}
\theta_{mf} = \arcsin \left( \frac{R}{len_{SP_f}} \right).
\end{equation}
The position of the tangent point M on the starting circle can be determined from the angle $\theta_M$ between the line segment $SM$ and the X-axis. And $\theta_M$ can be obtained from the geometric relationship with the angle $\theta_{sf}$ and the angle $\theta_{mf}$, as shown in the Table \ref{mytable2}.
\begin{table}[htbp]
    \begin{center}
    \caption{Geometric relationship of the determined angle $\theta_M$ of the tangent point $M$.}
    \label{mytable2}
    \begin{tabular}{lc}
        \toprule
        Path type & $\theta_M$\\
        \midrule
        LS & $\theta_{sf}+\theta_{mf}-\pi/2$ \\
        RS & $\theta_{sf}-\theta_{mf}+\pi/2$ \\
        \bottomrule
    \end{tabular}
    \end{center}
\end{table}
\begin{equation}
\label{myequation25}
(x_M, y_M) = \left( x_S + R \cos(\theta_M), y_S + R \sin(\theta_M) \right)
\end{equation}

Next, the starting arc $arc_s$ is calculated. This involves determining the angles of the starting arc $theta_s$. The angle $theta_s$ can be derived from its geometric relationship with the angle $\theta_M$ and the starting point attitude $\theta_s$, as illustrated in Table \ref{mytable3}. The length of the intermediate tangent segment, denoted as $len$, can be determined using the coordinates of the tangent points $M$ and $N$.
\begin{table}[htbp]
    \begin{center}
    \caption{Geometric relationship between the angles $theta_s$, $\theta_M$ and starting attitude $\theta_s$.}
    \label{mytable3}
    \begin{tabular}{lc}
        \toprule
        Path type & $theta_s$\\
        \midrule
        LS & $\theta_{M}+\theta_{s}+\pi/2$ \\
        RS & $\theta_{s}-\theta_{M}+\pi/2$ \\
        \bottomrule
    \end{tabular}
    \end{center}
\end{table}
\begin{equation}
\label{myequation26}
{len}_{arc_{s}} = R \cdot theta_{s}
\end{equation}
\begin{equation}
\label{myequation27}
{len}_{MP_f} = \sqrt{(y_f - y_M)^2 + (x_f - x_M)^2}
\end{equation}
Combining the two path planning problems, the path distance cost is given by the following equation:
\begin{equation}
\label{myequation28}
\begin{split}
L_{i, j}^{k} =
\begin{cases} 
len_{arcs} + len_{MN} + len_{arc_f}, & \text{CSC type}, \\
len_{arcs} + len_{Mf},  & \text{CS type}.
\end{cases}
\end{split}
\end{equation}
$L_{i,j}^{k}$ denotes the length of the Dubins path of the k-th fixed-wing UAV flying from the start point $i$ to the end point $j$. 
\renewcommand{\algorithmicrequire}{Input:}
\renewcommand{\algorithmicensure}{Output:}
\begin{algorithm}[h]
  \caption{: CS-type Dubins path generation}
  \label{myalgo1}
  \begin{algorithmic}[1]
    \Require
    $\boldsymbol{P_s}$,$\boldsymbol{P_f}$,$\boldsymbol{turn_s}$,$\boldsymbol{R}$
    \Ensure
      $\boldsymbol{arc_s}$,$\boldsymbol{len}$
    \While {Two CS-type path construction}
    \State Identify the turning circle
    \State Calculate turning circle center
    \State Calculate the endpoints of the tangent
    \State Obtain the path length
    \EndWhile
    \State Obtain the optimal path $ \leftarrow min\{RS, LS\}$
  \end{algorithmic}
\end{algorithm}

\section{REAL-TIME COLLABORATIVE MISSION PLANNING ALGORITHM BASED ON DUBINS PATH DISTANCE}
\subsection{Algorithm Overview and Characteristics}
To achieve real-time planning that adapts to dynamic and uncertain environments, this paper proposes an algorithm for distributed mission planning.
Decentralized decision-making significantly accelerates decision-making process while enhancing the system's robustness. 
In the distributed approach, each UAV functions as an independent intelligent agent that ensures task space consistency by sharing and negotiating their respective resource and task demand information, ultimately leading to an overall reasonable task allocation scheme. 

The distinguishing feature of the algorithm is its rejection of cumbersome iterative processes, retaining  essential low-level iterations with minimal computational overhead. 
The optimal path typically results from iteration; however, such iterations often entail a heavy computational burden. 
The goal of the algorithm is not to achieve the optimal allocation in a computationally light iteration, but rather to obtain a near-optimal or acceptable allocation. This is trade-off between optimality and real-time. 
Therefore, we focus on allocation strategy with low computational effort: the hungarian algorithm in deterministic methods and the greedy algorithm in heuristic approaches. The greedy algorithm is straightforward and does not require information from other UAVs for decision-making. While the Hungarian algorithm requires position information of other UAVs for decision-making. However, it can theoretically yield the optimal solution for a single task assignment. 

A further strength of the algorithm is that instead of utilizing the conventional Euclidean distance between two task points, it employs the Dubins path distance as a cost function. Utilizing the Dubins path distance offers three benefits. 
Firstly, the Dubins path is the flyable path that a fixed-wing UAV can follow between two task points, as it satisfies the UAV’s kinematic constraints.
Therefore, the Dubins path length accurately characterizes the actual kinematic distance between the two task points. 
Decisions based on this distance cost may differ from those applying straight-line distances, leading to variations in mission planning scheme.
Secondly, the proposed Dubins path generation method quickly obtains the paths and computes the distance cost to meet the time constraints.
This is made possible by reducing the types of Dubins path solutions and simplifying the form of the solutions.
Finally, connected paths, both with and without terminal constraints, can be planned using Dubins paths. This in turn makes it possible to fully take into account the heading angle constraints during the task assignment phase. Further, this allows our algorithm to handle heterogeneous tasks.

Before all UAVs make decisions, the algorithm starts with the clustering pre-processing for all the tasks.
This is another distinguishing feature of the algorithm. The purpose of clustering is twofold. 
First, task clustering allows different UAVs to be assigned to more centralized task clusters. The paths between the UAVs are almost non-intersecting or slightly overlapping. 
Thus, the total distance travelled by all UAVs is reduced, and the task allocation tends to be as optimal as possible.
Then, the size of the decision space directly impacts the duration of the UAV decision-making process. 
It takes less time to obtain a rational allocation scheme using some of the tasks as the decision space than all of the tasks as the decision space.
Clustering preprocessing can effectively partition the original large decision space into smaller subsets allocated to each UAV. 

\begin{algorithm}[ht]
  \caption{: Multi-UAV real-time collaborative mission planning}
  \label{myalgo2}
  \begin{algorithmic}[1]
  \State Initialization. Total number of drones $K$, total number of missions $N$ , range of motion ROM, drone configuration $P$, the UAV state $U_S$, state of the mission $T_S$
  \State $TaskPosition=ROM*rand(N,2)$
  \State Obtain reasonable results for task clustering
  \While {All task unfinished}
  \State Get the current status of each UAV $U_s$
  \If {UAV status is idle}
   \State Check the unfinished tasks $T_{(Na_list)}^k$
   \State Get current UAV configuration $p$
   \State Initialize the cost matrix $cMat$
   \For {Each unassigned mission}
   \State Get the location of the mission point
   \State Construct paths and obtain lengths $L_{ij}^k$
   \State Place the path length into the cost matrix
   \EndFor
   \State Reconstruct the matrix of Costs $cMat'$
   \State Fast access to allocation schemes 
   \If {UAV assigned to a mission}
   \State Change the status of the task 
   \State UAV status changes to in-transit
   \State UAV receives path and begins to move
   \EndIf
  \ElsIf {UAV status is in-transit}
  \State Distance to mission point $dvec$
  \If {Distance less than arrival criterion}
  \State The UAV arrives at the mission point 
  \State The status is changed to busy
  \Else
  \State The UAV continues to fly 
  \EndIf
  \Else  { UAV status is busy}
  \State Determine the task type and plan path
  \If{Mission finished criteria are met}
  \State Mission status changes to completed 
  \State UAV status changes to idle
  \Else
  \State UAV continues to fly coverage path
  \EndIf
  \EndIf
  \EndWhile
  \State UAV swarm return
  \end{algorithmic}
\end{algorithm}

\subsection{Procedure of Algorithm}
\begin{figure*}[ht]
\centering
\includegraphics[width=0.55\linewidth]{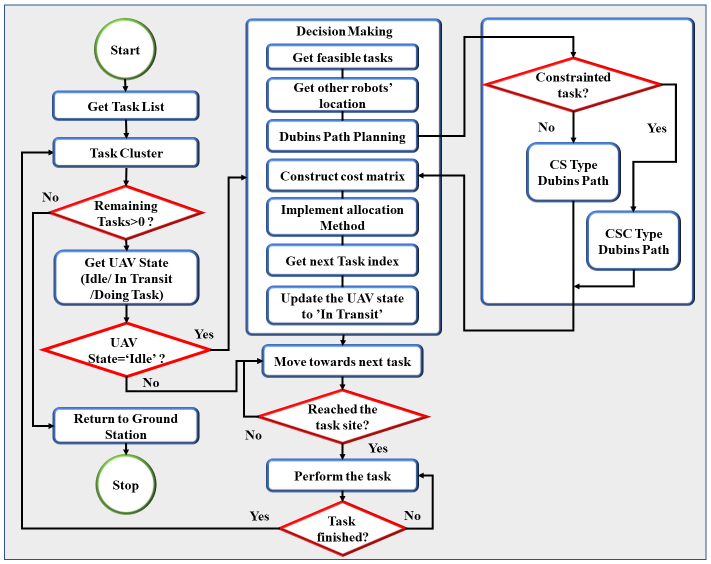}
\caption{Multi-UAV real-time collaborative mission planning flowchart.}
\label{myfigure6}
\end{figure*}
Dynamic real-time multi-UAV collaborative mission planning algorithm based on Dubins paths is proposed. 
The pseudo-codes of algorithm are shown in Algorithm 2 and the procedures are detailed as follows.

\textbf{Step 1.} Initialization (lines 1-2). Given the UAV's prior parameters, such as the total number, configuration, and state, along with the range of motion and the total number of tasks. A random function is then employed to generate the position of each task point, resulting in all tasks being initially unassigned. 

\textbf{Step 2.} Task Clustering (line 3). The K-means algorithm is employed to classify all tasks into categories matching the total number of UAVs and to assign them to the corresponding UAVs based on appropriate rules. 

\textbf{Step 3.} Assessment of Task Completion (line 4). The status of all tasks are assessed. Based on the completion status of all tasks, it is determined whether the overall tasks of the UAV swarm are completed and the UAV swarm needs to return to the base.

\textbf{Step 4.} Obtain the UAV State (line 5). The state of each UAV serves as the basis for determining its next movement, and the three distinct UAV states exhibit different movement patterns.

\textbf{Step 5.} Decision-making in the Idle State (lines 6-21). 
The UAV first forms a list of unassigned tasks and obtains the configuration of the UAV. Then, the cost matrix is initialized and the distance cost is obtained using the path generator. Next, the appropriate assignment algorithm is invoked to obtain the assignment scheme. State changes are made based on whether the UAV is assigned to a task or not.

\textbf{Step 6.} Decision-making in the In-transit State (lines 22-29). 
Firstly, the distance between the UAV and the target point at the current time is calculated, and the arrival criterion is used to determine whether the UAV has reached the mission point. When the UAV reaches the task point, the UAV's state at this time should be doing the task; conversely, the UAV should continue to fly the planned path.

\textbf{Step 7.} Decision-making in the Busy State (lines 30-37).
Firstly, the UAV needs to plan the coverage path of the mission in real time based on the type of mission. Then the UAV executes the task and flies the coverage path. Finally, when the task completion criteria are met, the UAV completes the relevant task and its state is changed to an idle state.


\textbf{Step 8.} UAV Return (line 40). Once all tasks are completed, the UAV swarm can return. Each UAV utilizes the path planner to plan its path from the current position back to the base in real time and then begins flying autonomously.

\section{EMERGENCY SITUATION}
The vulnerability of UAVs and external environments to sudden changes necessitates that UAV swarms possess real-time planning capabilities to provide effective schemes that satisfy numerous constraints during emergencies. Real-time mission planning and re-planning during emergencies further emphasize the time constraints on algorithms, necessitating that they complete mission planning within limited computational resources and time. 

\subsection{Strategy for New Tasks}
The most common emergent scenario for UAVs during the flight is the emergence of new tasks. 
The foundational parameters of the new scenario are similar to those of a static scenario, including UAV performance, mission attributes, and so on.
The emergence of new tasks implies real-time planning, including task clustering and real-time task allocation.
Therefore, the challenges of addressing this emergent situation are twofold.
The proposed method is a real-time planning approach that does not rely on a predetermined task sequence. Therefore, it is inherently capable of addressing the real-time planning. 
Additionally, the emergence of new tasks necessitates preprocessing for clustering. 
This tricky problem is the focus of our solution efforts.

Emergencies are the triggering conditions for the dynamic planning problem. An indeterminate number $s_1$ of new tasks $T^D$ can emerge at any time during the flight. 
Let $\bar{T}$ denote the set of new tasks that need to be assigned, $\bar{T} = \left\{ T_{N+1}^{D}, T_{N+2}^{D}, \ldots, T_{N+s_{1}}^{D} \right\}$ where $T_{N+i}^{D} \left(i \in \left\{1,2,\ldots, s_{1}\right\}\right)$ denotes the $N+i-th$ new task to be assigned. 
However, for the sake of simplicity, this paper assumes that new tasks appear randomly within a specific time window. 
And the new tasks are non-urgent and can be completed by UAVs without time constraints during the mission. It is also assumed that the emerging tasks are simple tasks, i.e., point targets, and that any UAV is capable of performing the task. 
To process new tasks, the UAV formation must first gather relevant information $T_{triple}$ about the new tasks, including its location and type, and then share information within the formation.
The new tasks still require preprocessing through clustering, but in a different way than the initial static tasks. Since the clustering of the initial tasks has established an acceptable criterion, the criterion for the new tasks should align with it. When the initial tasks are clustered, centroids for clusters $x_{cluster_center}^i$ with an equal number of UAVs are generated. The new tasks first compute the Euclidean distance to each cluster centroid,
\begin{equation}
\label{myequation29}
dis^{i} = \left\| x^{D} - x_{\text{cluster\_center}}^{i} \right\|_{2}.
\end{equation}
Then, based on the distance $dis^i$, the new tasks are categorized into a specific task cluster and assigned to a particular UAV. The task cluster index is calculated as follows:
\begin{equation}
\label{myequation30}
i = \operatorname{argmin}(dis^{i}).
\end{equation}

After a new task is assigned to a UAV, two possible situations arise regarding its execution. In the first case, the UAV has not fully completed the initial tasks, and some tasks remain. Thus, adding new tasks may alter the original task execution sequence, including the order and connection paths of the tasks. In the second scenario, the UAV has completely accomplished the initial tasks. When new tasks appears, the UAV immediately proceeds to the new tasks without altering the original task execution sequence. 
\begin{algorithm}
    \caption{: New tasks generated and categorized}
    \label{alg3}

    \begin{algorithmic}[1]
    \State $probofocc \leftarrow rand\quad function$
    \If {$probofocc < Occurrence \quad criteria$}
      \State $x^D=ROM*rand(N,2),T_{state}^D=0,T_{type}^D$
      \State total task number $N+1$
    \EndIf
    \For {each cluster centroid $i$}
      \State $dis^{i} = \left\| x^{D} - x_{\text{cluster\_center}}^{i} \right\|_{2}.$
    \EndFor
    \State $i = \operatorname{argmin}(dis^{i})$
    \end{algorithmic}   
\end{algorithm}

\subsection{Strategies for UAV Damage} 
Due to mechanical failures and communication breakdowns, the UAVs are not functioning properly and, naturally, can no longer carry out their tasks.
Then the UAVs are deemed to be in a damaged state. 
Obviously, damaged UAVs reduce the number of available UAVs in the UAV swarm and also alter the original mission planning scheme. 
This indicates that UAV swarms should undergo mission replanning.

Let $\overline{U}$ denotes the set of UAVs that can perform the mission, $\overline{U} = \{U_1, U_2, \ldots, U_{K-S_2}\} = U - \{U_1, U_2, \ldots, U_{S_2}\}$ where $s_2$ denotes the number of damaged UAVs and $U_j (j \in \{1, 2, \ldots, s_2\})$ denotes the $j-th$ damaged UAV.
At the time of damage, the UAV may not have completed all of its assigned tasks. This can cause significant problems in real-time task planning.
Therefore, a categorized discussion of the questions based on task completion follows.

UAV damage does not affect mission planning when all missions assigned to that UAV are completed, it simply makes fewer UAVs available.
In contrast, the situation becomes complicated when there are outstanding tasks, as the UAV may be in one of two states: in-transit or busy.
In either case, the UAV has already been assigned a specific task, so that task is in the assigned state.
Sudden damage to the UAV directly prevents the task from continuing, causing the UAV immediately releases the task, and the task status becomes unassigned.
The other unfinished tasks originally assigned to the UAV are reassigned to other available UAVs according to certain rules.
Tasks are reassigned to different UAVs based on their distance from the cluster centres of other available UAVs and according to the proximity principle.



\begin{algorithm}
    \caption{: UAV damage response strategies}
    \label{alg4}

    \begin{algorithmic}[1]
    \State $damagetime =  50+10*rand$
    \If {$runtime = damagetime$}
      \State  $k=failindex = randi(K), U_{state}^k=3$
      \If {UAV is not idle}
       \State  Mission release received by UAV,$T_{state}^k=0$
       \For{each task in cluster}
         \If {task not completed}
           \For {available UAVs $i$}
              \State $dis^{i} = \left\| x^{D} - x_{\text{cluster\_center}}^{i} \right\|_{2}.$
           \EndFor
           \State $i = \operatorname{argmin}(dis^{i})$
         \EndIf
       \EndFor
      \EndIf
    \ElsIf {$runtime > damagetime$}
      \State $U_{state}^k=3$, UAV stops moving
    \Else 
      \State All UAVs are operating normally
    \EndIf
    \end{algorithmic}   
\end{algorithm}

\subsection{Emergencies Occurring Simultaneously}
\begin{figure}[h]
\centering
\includegraphics[width=0.85\linewidth]{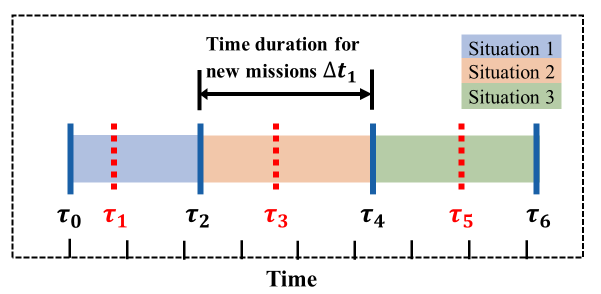}
\caption{ The sequential relationship between the destruction time of drones and the emergence of new tasks is as follows: $\tau_0$ and $\tau_6$ represent the start and end times of the UAV swarm mission, respectively. The time interval $ [\tau_2, \tau_4 ]$ indicates the window during which a new mission may arise, while $\tau_1$, $\tau_3$, and $\tau_5$ correspond to the times at which UAVs are destroyed in each of the three scenarios.}
\label{myfigure7}
\end{figure}

Extreme situations where new tasks and UAV destruction occur simultaneously are complicated. To analyze the various scenarios in a comprehensive and clear manner, we divided them into three categories based on the sequential relationship between the timing of UAV destruction and the emergence of new missions, as shown in Fig. \ref{myfigure7}.

1) UAV damage occurs before new tasks arise, $\tau_1 < \tau_2$. The entire process of mission planning in this scenario involves two phases. First, the mission planning under UAV damage takes place. After the UAV is destroyed, the new tasks emerge. Then, the scenario can be considered a new mission planning under  with a reduced number of available UAVs when new tasks arise.

2) UAV destruction occurs after the new tasks are no longer active, $\tau_4 < \tau_5$. The complete mission planning process in this scenario consists of two phases. First, the UAV swarm conducts mission planning with only new tasks emerging. Subsequently, the new task ceases to appear, and the UAV is destroyed. 
This is equivalent to a mission planning with an increase in total mission tasks when UAV damages.

3) UAV destruction during the new mission emergence period, $\tau_2 < \tau_3 < \tau_4$. The complete mission planning process in this scenario consists of three distinct phases. First, the UAV swarm conducts mission planning with only new tasks appearing. Suddenly, the destruction situation occurs. And the swarm begins equivalent mission planning for UAV damage with an increase number of tasks. Subsequently, new tasks continue to emerge. The swarm engages in mission planning for the emergence of new missions with a reduced number of available UAVs. 

In summary, although the simultaneous occurrence of two emergencies generates numerous possible scenarios for mission planning, identifying a pattern becomes straightforward upon careful analysis. The emergence of a new mission merely increases the total number of missions, which does not significantly affect the replanning of swarm in the event of UAV damage. 
On the contrary, UAV destruction leads to a reduction in available UAVs, directly impacting the preprocessing of clustering for new tasks.
Therefore, when two emergencies occur simultaneously, only the part of the clustering processing of the new tasks need to be adjusted to accommodate the full scenario. The adjustment involves determining whether the corresponding UAV is available before calculating the distance between the new task and the clustering center.

\section{SIMULATION RESULT AND ANALYSIS}
\subsection{Simulation Environment Settings}
These algorithms were deployed and tested in MATLAB R2022a on a standard PC running Windows 10, equipped with an Intel Core i7 2.2 GHz CPU and 8GB of RAM. 

The nominal speed of the UAV varies depending on the size of the mission area in different scenarios. However, all UAVs in the same scenario share the same nominal speed. The minimum turning radius of the UAVs is 80m. The UAVs are fixed-wing aircraft with a large aspect ratio and long endurance. The size of the mission area ranges from 2km×2km to 3km×3km. Different mission tables provide information about the missions in various scenarios, including the number of missions, their locations, and types. 
The UAVs depart from the base $S$ simultaneously to reach their respective mission points, complete the missions, and then return to the base after finishing the tasks. It should be noted that all UAVs return to the base station together only when all the tasks are completed. This implies that some UAVs may enter a hovering or waiting state at the final stage of the flight.
The number of UAVs varies depending on the mission scenario, with most scenarios involving four UAVs. 

To validate the effectiveness of the proposed algorithmic framework, it is compared with the simulated annealing algorithm (SA), the greedy-based task assignment approach with Euclidean distance (GBA), the hungarian-based approach with Euclidean distance (HBA), the auction-based approach with Euclidean distance (AA). 
Where SA is used as a benchmark with an initial temperature of 50º and a hyperparameter value of 0.99 for exponential cooling. Each Markov chain is run 500 times and the algorithm is terminated when the temperature drops below 10º or the total number of iterations reach 1000.
Specifically, the application of the proposed framework leads to the preprocessing-enabled real-time mission planning approach based on Dubins path distance (PRBDD) and its variant without preprocessing operations (RBDD).
These two approaches are abbreviated as RBDDG and PRBDDG in conjunction with the adopted greedy task allocation strategy.
In the subsequent sections, the six algorithms mentioned above are compared through simulations.
Note that we applies Dubins paths to smooth the straight paths obtained by SA, GBA, HBA and AA to obtain the final path lengths.

\subsection{Deterministic Simulation}
\subsubsection{Cost Function Performance}
The efficiency of calculating the distance cost function directly determines the speed of mission planning. 
Fast and accurate calculation of the distance between the UAV's current position and each task to be assigned is a crucial cornerstone of task assignment. 
Therefore, we quantitatively assess the computational efficiency of the distance cost function applied in the proposed algorithm. 
The experiment is divided into three groups, control group (the Euclidean distance group), CS-type path group and CSC-type path group, and the simulation is repeated 50 times.
The experiment randomly generates 50 point pairs each. 
Each point pair consists of a start point and an end point, with the positions preset as mandatory information, while the heading angle information is provided as needed.
The three groups of point pairs share identical position information, but differ in their heading angle.


\begin{table}[htbp]
    \begin{center}
    \caption{Calculation time required for different distance costs.}
    \label{mytable4}
    \begin{tabular}{lc}
        \toprule
        Distance cost function & Computing time (s)\\
        \midrule
        Euclidean distance & 0.00040 \\
        CS type Dubins distance & 0.00053 \\
        CSC type Dubins distance & 0.00117\\
        \bottomrule
    \end{tabular}
    \end{center}
\end{table}

Table \ref{mytable4} shows the average computation time for the three distance cost functions. The Euclidean distance has the shortest computation time, which is consistent with our intuition. The CS type Dubins path distance requires slightly more computation time than the Euclidean distance. However, they both fall within the same order of magnitude, at approximately $10^{-4}s$. The CSC type Dubins path distance has the longest computation time of the three, on the order of milliseconds. But it's also faster than the $0.058s$ of the current state-of-the-art algorithms. 
The additional computational speedup of the $Dubins\_{CS}$ function, by one more order of magnitude, results from a further reduction in solution types and a simpler solution structure.

\subsubsection{Allocation Based on Different Distance Cost Functions}
\begin{figure}[ht]
\centering
\subfigure[]{
\includegraphics[width=0.7\linewidth]{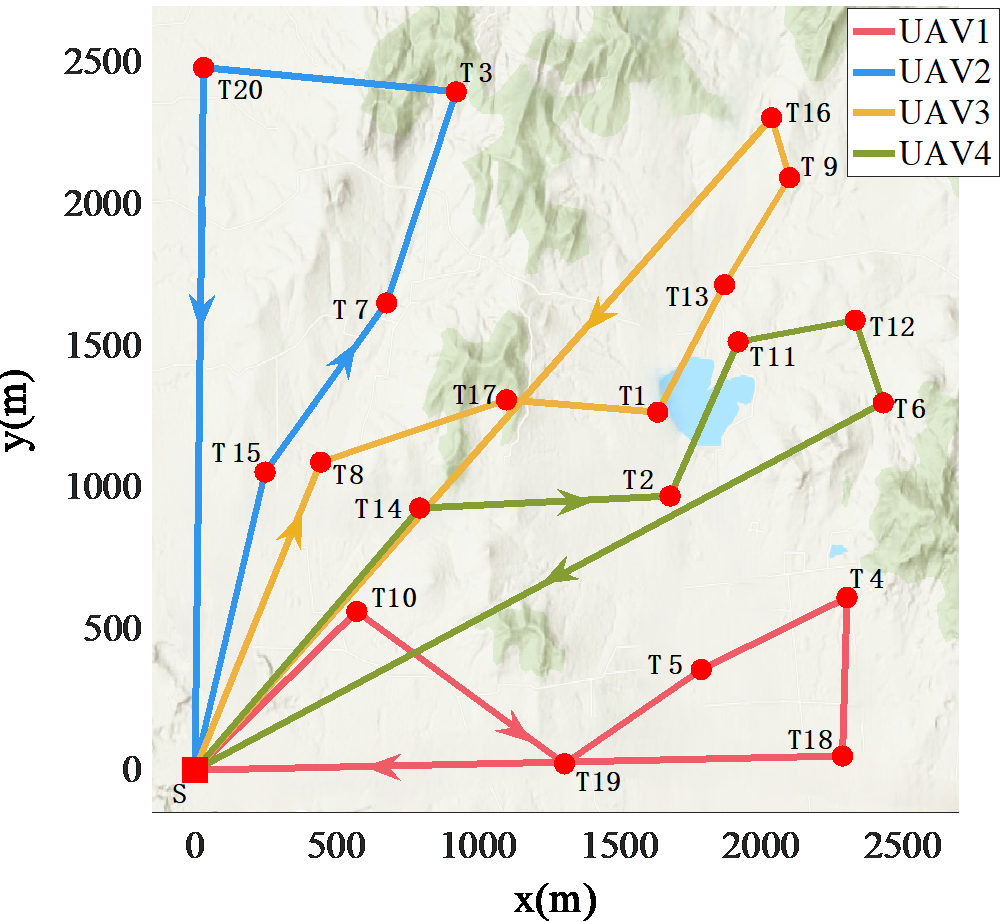}
\label{myfigure8}
}
\subfigure[]{
\includegraphics[width=0.7\linewidth]{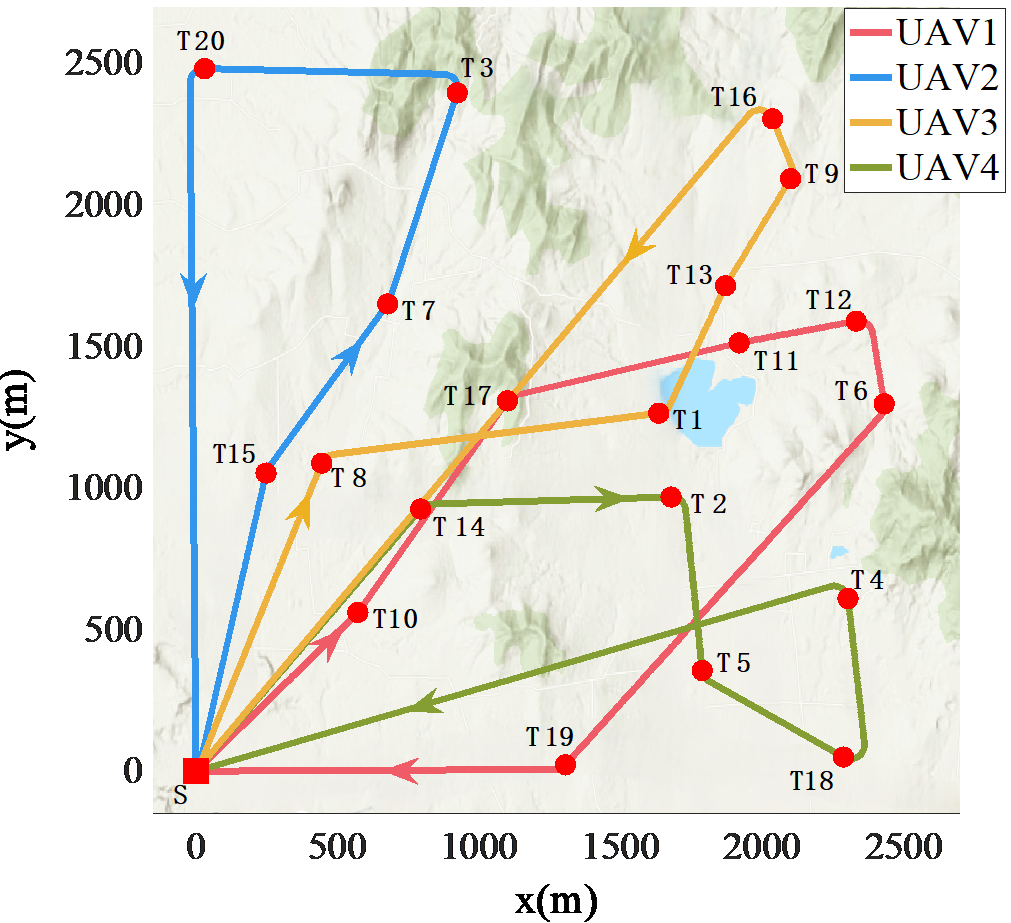}
\label{myfigure9}
}

\caption{(a) Mission planning based on Euclidean distance. (b) Mission planning based on Dubins path distance.}
\end{figure}

Differences in the distance cost function employed for task assignment may lead to significantly different assignment results, as well as affect the assignment time. This claim is validated in the following mission scenario. 
The scenario involves four homogeneous UAVs completing 20 tasks evenly distributed over a 2.5 km x 2.5 km area.
It is assumed that all UAVs fly at a constant speed throughout the mission, with a speed set to $17.5m/s$.
Since the Euclidean distance cost function cannot take into account the heading angle constraints, the task is limited to point task without heading angle constraint, which can be completed by any UAV. The locations of the tasks are randomly generated, and their distribution within the task area is shown in Figs. \ref{myfigure8}-\ref{myfigure9}. The red square indicates the base station coordinates $(0,0)$, and the red dot represents the task locations. 
The methods applied in this scenario are GBA and RBDDG.
The difference between Euclidean distance task assignment and Dubins path distance assignment lies in the fact that one uses Euclidean distance, while the other uses Dubins path distance to calculate the distance cost function. 

As can be seen from Fig. \ref{myfigure8}, the mission planning based on Euclidean distance only completes the task assignment process, while path planning is not performed.
The resulting path is a straight line, ignoring the heading angle, which prevents the UAV from following the path directly. 
In contrast, the path generated by Dubins-based mission planning is continuous and smooth, aligning with the UAV's kinematic constraints, as in Fig. \ref{myfigure9}.
In addition, the two planning methods exhibit entirely different distribution schemes.
Except for UAV 2, which has the same assignment, the other three UAVs not only have different tasks in the different mission schemes, but also the order in which the tasks are performed is different, as Table \ref{mytable5}.
Due to the differences in both the tasks and their execution order, the distances flown by UAVs also vary across the different methods.
Comparative experiment demonstrates that the difference in distance cost has an impact on the task assignment results.


\begin{table}[htbp]
    \begin{center}
    \caption{Mission planning details based on different distance costs.}
    \label{mytable5}
    \begin{tabular}{lcc}
        \toprule
        \makecell[c]{UAV \\ number} & \makecell[c]{Dubins distance \\task allocation} &  \makecell[c]{Euclidean distance \\task allocation}\\
        \midrule
        1 & \makecell[c]{6: T10 \textcolor{red}{→T17}\\\textcolor{red}{→T11→T12}\\\textcolor{red}{→T6→T19}} &  \makecell[c]{5: T10 \textcolor{red}{→T19}\\ \textcolor{red}{→T5→T4}\\\textcolor{red}{→T18}}\\
        2 & \makecell[c]{4: T15→T7\\→T3→T20} & \makecell[c]{4: T15→T7\\→T3→T20}\\
        3 & \makecell[c]{5: T8→T1\\→T13→T9\\→16} & 
        \makecell[c]{6: T8 \textcolor{red}{→T17}\\→T1→T13\\→T9→T16}\\
        4 & \makecell[c]{5: T14→T2 \\ \textcolor{red}{→T5→T15} \\ \textcolor{red}{→T4}} & \makecell[c]{5: T14→T2 \\ \textcolor{red}{→T11→T12} \\ \textcolor{red}{→T6}}\\
         \bottomrule
    \end{tabular}
    \end{center}
\end{table}


To further illustrate how different distance costs affect mission planning, we take a closer look at Figs. \ref{myfigure8}-\ref{myfigure9}.
The key to all subsequent task assignments is that UAV 1 chooses task 17 instead of task 19.
To better understand the logic behind decision-making, the Dubins path distance is divided into two components: the straight-line distance between the previous and next task points, and the additional distance due to heading angle correction.
In Euclidean distance based mission planning, task 19 is closer to task 10 in a straight distance, resulting in the UAV selecting task 19 as its next task.
In contrast, in Dubins path distance-based mission planning, the additional heading angle correction distance for task 19 is significantly longer than that for task 17, increasing the total path length and leading the UAV to select task 17.
This is the mechanism by which the additional distance due to heading angle correction affects task assignment. 
The key prerequisites for its effectiveness are that the two tasks are at similar straight-line distances from the current position and that their heading angles differ.


\subsubsection{Comparison and Evaluation of Proposed Algorithms}
\begin{figure}[ht]
\centering
\includegraphics[width=0.85\linewidth]{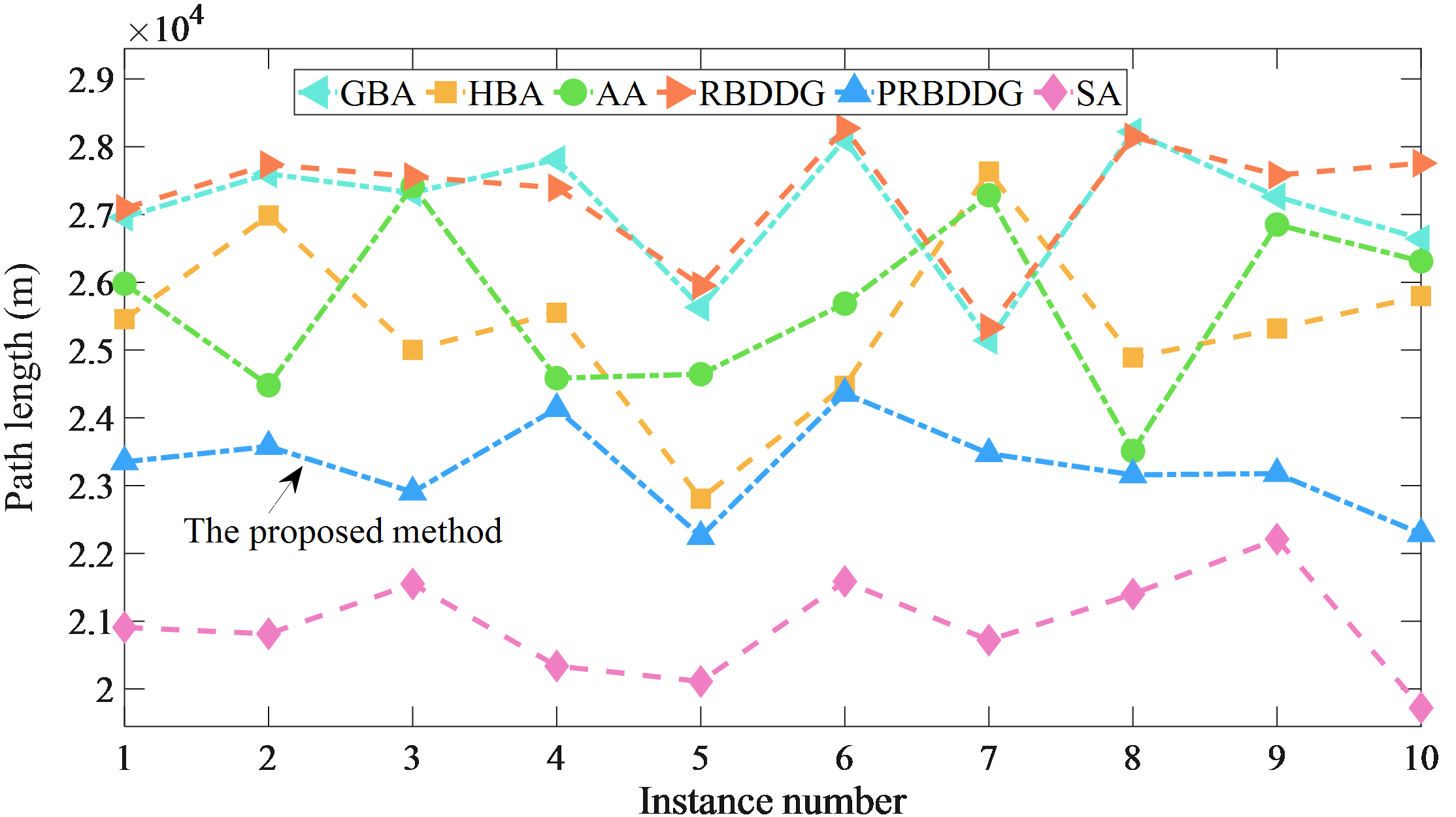}
\caption{Spatial comparison results of 6 different methods: the total distance.}
\label{myfigure10}
\end{figure}

To fully characterize the proposed algorithms, this section compares the six algorithms in both spatial and temporal dimensions. The experimental scenario in this section is the same as that in the previous section.
The simulation is conducted 50 times, each time with 25 tasks randomly and uniformly distributed without repetition.
Fig. \ref{myfigure10} shows the total distance traveled by four UAVs applying six methods in 10 out of 50 simulation results.
Obviously, SA obtains the optimal allocation scheme, resulting in the shortest distance traveled by the UAV. 
As shown in the figure, all four methods, RBDDG, GBA, HBA and AA, produce only feasible allocation schemes, with their total UAV travel distance being significantly longer than the optimal solution, which is close to each other. 
On the contrary, PRBDDG obtains suboptimal solutions or even very close to optimal solutions.
The gap refers to the difference between the total distance traveled by the UAVs in the other methods and that obtained by the SA, with a smaller gap indicating a method that is closer to the optimal solution. The average gap for PRBDDG is $9.57\% $, respectively, while the average gaps for the other four methods are all above $20\%$. 
The comparison confirms the proposed methods obtain a suboptimal allocation.

\begin{figure}[ht]
\centering
\includegraphics[width=0.85\linewidth]{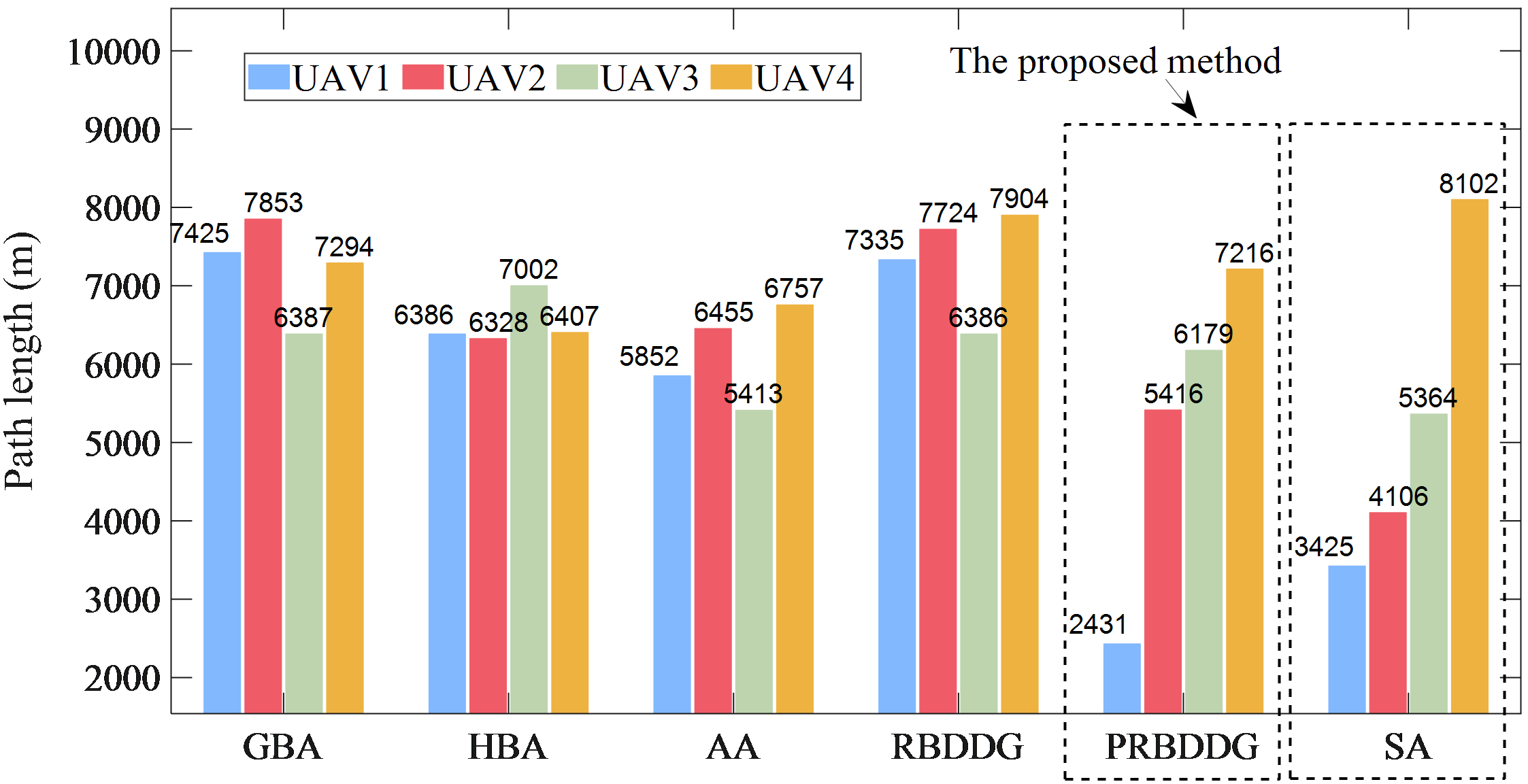}
\caption{Spatial comparison results of 6 different methods: distance traveled by each UAV in the 4th example. The SA method marked by the dashed line is the optimal solution. The other dashed line marks our proposed methods.}
\label{myfigure11}
\end{figure}

\begin{table}[htbp]
    \begin{center}
    \caption{Spatial correlation quantities of 6 different methods.}
    \label{mytable7}
    \begin{tabular}{lccc}
        \toprule
          Mehthod & \makecell[c]{Average \\ total \\distance (m)} & \makecell[c]{Maximum  \\ distance \\difference (m)} &  \makecell[c]{Maximum \\ task number \\difference}\\
        \midrule
        GBA & 27070.04 & 1364.47 & 2\\
        HBA & 25491.05 & 1292.14 & 2\\
        AA & 25675.34 & 1577.99 & 2\\
        RBDDG & 27282.81 & 1397.68 & 3\\
        PRBDDG & 23268.14 & 3895.62 & 6\\
        SA & 21235.73 & 4755.36 & 8\\
        \bottomrule
    \end{tabular}
    \end{center}
\end{table}

Fig. \ref{myfigure11} displays the distance traveled by each of the four UAVs employing the six different methods in the fourth example. As shown in the figure, the distances traveled by each UAV for PRBDDG are similar to those for the SA, with some UAVs traveling farther and others traveling shorter distances. In contrast, the distances for the remaining four methods are more average allocation. 
The reason for this phenomenon is the clustering of tasks. The PRBDDG effectively avoid task assignment averaging by applying clustering preprocessing to assign tasks differentially to UAVs.
To further illustrate that differential allocation leads to optimal solutions, we also calculated the difference between the longest and shortest distances traveled by the four UAVs, as well as the difference in the number of missions assigned to the UAVs. As shown in Table \ref{mytable7}, the difference in distance traveled and the difference in the number of missions for the SA are the largest among the six methods, followed by PRBDDG, with the smallest differences seen in the other four methods. 
This indicates that the optimal solution involves differentially assigning tasks, rather than distributing tasks equally among UAVs. 
Differential allocation can effectively avoid crossing or overlapping paths, thereby shortening the total path.
This is the mechanism that allows PRBDDG to achieve near-optimal suboptimal solutions.

\begin{figure}[ht]
\centering
\includegraphics[width=0.85\linewidth]{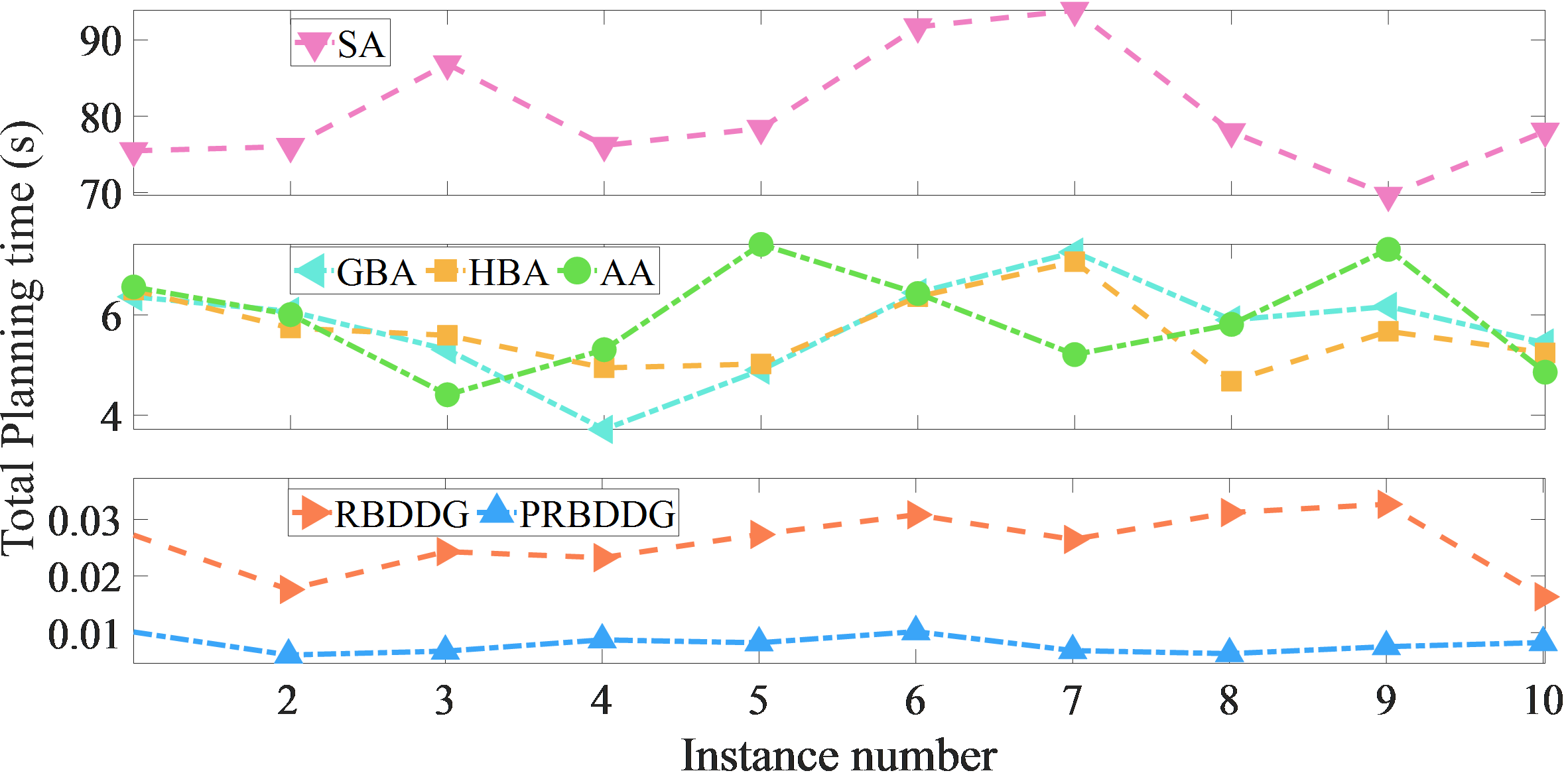}
\caption{Time comparison results of 6 different methods: total time for task planning.}
\label{myfigure12}
\end{figure}

\begin{figure}[ht]
\centering
\includegraphics[width=0.8\linewidth]{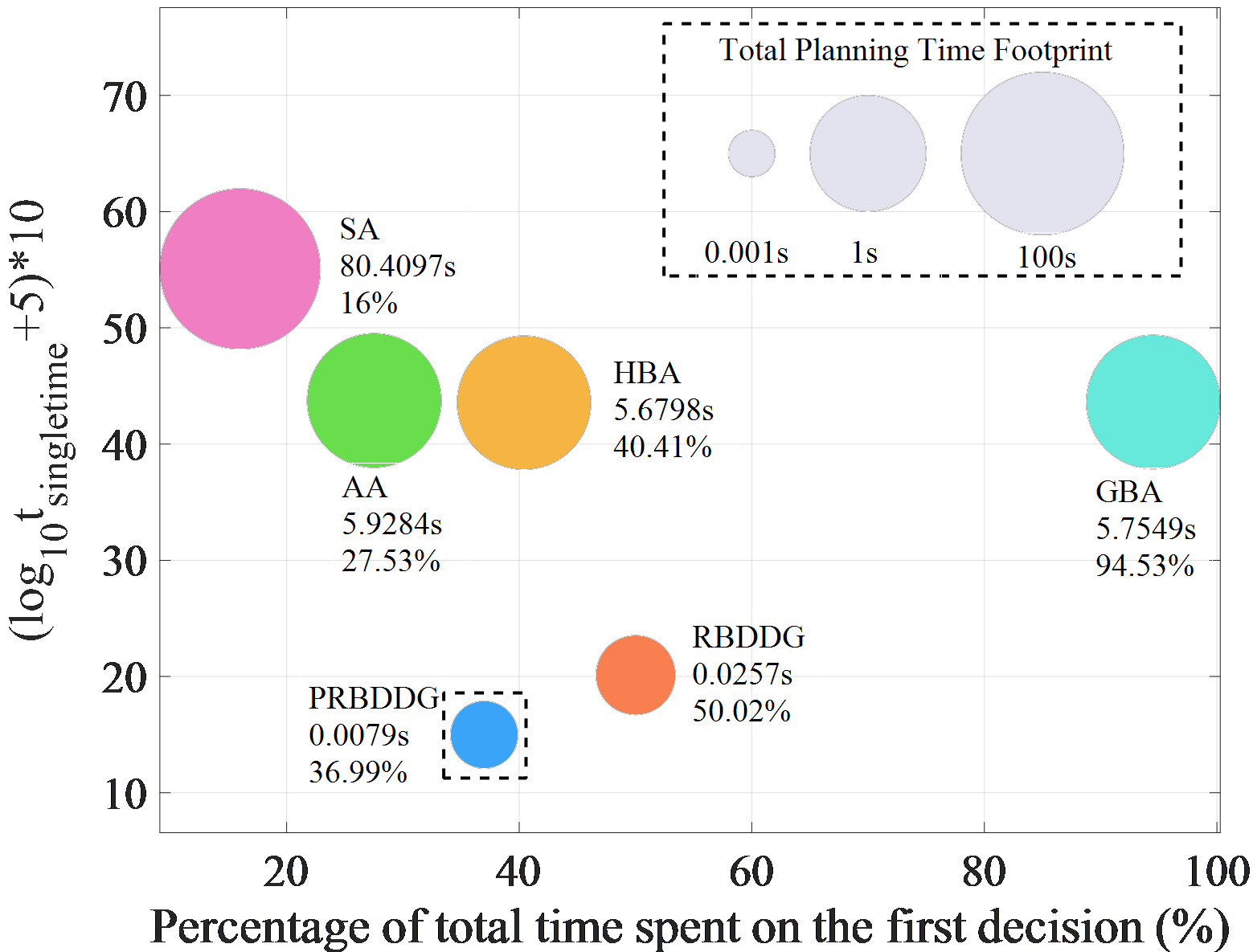}
\caption{Time comparison results of 6 different methods. }
\label{myfigure13}
\end{figure}

The comparison of the methods in terms of time is shown in Figs. \ref{myfigure12}-\ref{myfigure13}.
Fig. \ref{myfigure12} displays the total mission planning time by four UAVs applying six methods in 10 out of 50 simulation results.
Obviously, PRBDDG has the shortest total mission planning time. 
The RBDDG is next best, with total times ranging from a few milliseconds to tens of milliseconds. 
This suggests that clustering preprocessing further reduces the time required for mission planning.  
The total mission planning time for GBA, HBA, and AA is in the range of a few seconds, while the total time for the SA approach is measured in minutes.

\begin{table}[htbp]
    \begin{center}
    \caption{ Time related variables for 6 different methods.}
    \label{mytable8}
    \begin{tabular}{lccc}
        \toprule
          Mehthod & \makecell[c]{Average \\ total \\planning time (s)} & \makecell[c]{Average  \\ planning \\time (s)} &  \makecell[c]{Percentage of \\ total time  \\spent on the\\ first decision (\%)}\\
         \midrule
        GBA & 5.7549 & 0.2302 & 94.53 \\ 
        HBA &  5.6798 & 0.2272 & 40.41\\
        AA & 5.9284 & 0.2371 & 27.53\\
        RBDDG & 0.0257 & 0.0010 & 50.02\\
        PRBDDG & 0.0079 & 0.0003 & 36.99\\
        SA & 80.4097 & 3.2164 & 16 \\
        \bottomrule
    \end{tabular}
    \end{center}
\end{table}

Fig. \ref{myfigure13} exhibits the performance of the seven methods in terms of average total planning time, average single planning time, and time spent on the first decision as a percentage of the total time.
The size of the circles in different colours indicates the average total planning time required; the larger the circle, the longer the time needed.
It can be seen that PRBDDG and PRBDDH require the shortest average total planning time, followed by RBDDG and RBDDH, with SA requiring the longest time.
The average planning time for different methods varies by orders of magnitude, so showing all methods on the same graph requires mathematical manipulation. The approach is to take logarithms before scaling.
The vertical axis of Fig. \ref{myfigure13} then represents the average single planning time, with larger values implying longer times.
The results of the comparison of the average single time for the seven methods are consistent with the comparison of the total time.
The results also suggests clustering preprocessing is also effective in reducing the single mission planning time by decreasing the size of the decision space.
It is worth noting that the single planning times for PRBDDG and PRBDDH are on the order of $10^{-4}$, as shown in the Table \ref{mytable8}.
This is consistent with the time scale discussed in the previous section.
The horizontal axis of Fig. \ref{myfigure13} represents the proportion of the total time taken by the UAV to complete the first mission planning while hovering at the starting point.
A higher percentage in this ratio indicates a shorter mission planning time during flight.
The GBA algorithm has the highest time percentage, while PRBDDG and PRBDDH have a small percentage.
The proposed PRBDDG/PRBDDH algorithms are close in performance in both spatial and temporal dimensions and can be used with each other in most scenarios. However, the PRBDDG algorithm has a small advantage in terms of computational efficiency.

Based on the quantitative comparisons above, Table \ref{mytable100} summarizes the key features of each algorithm, providing a more intuitive reflection of the strengths and weaknesses of each algorithm.
\begin{table}[h]
    \begin{center}
    \caption{Key performances comparison of each mission planning method: Coupled solution(CS), Real time performance (RT), Emergency (EM), Path (PA). An asterisk denotes our proposed algorithm. \ding{52} indicates that the method performs well in terms of performance, while \ding{56} indicates that the method performs poorly in terms of perfor-mance improvement.}
    \label{mytable100}
    \begin{tabular}{lcccc}
        \toprule
        Methods & CS  & RT & EM & PA \\
        \midrule
        SA & \ding{56} & Bad & \ding{56} & Optimal \\
        GBA / HBA / AA & \ding{56} & Fair & \ding{56} & Nonoptimal \\
        RBDDG & \ding{52} & Good & \ding{52} & Nonoptimal \\
        PRBDDG & \ding{52} & Excellent & \ding{52} & Suboptimal \\
        \bottomrule
    \end{tabular}
    \end{center}
\end{table}

\subsubsection{Typical Application Scenario}
\begin{figure}[ht]
\centering
\includegraphics[width=0.8\linewidth]{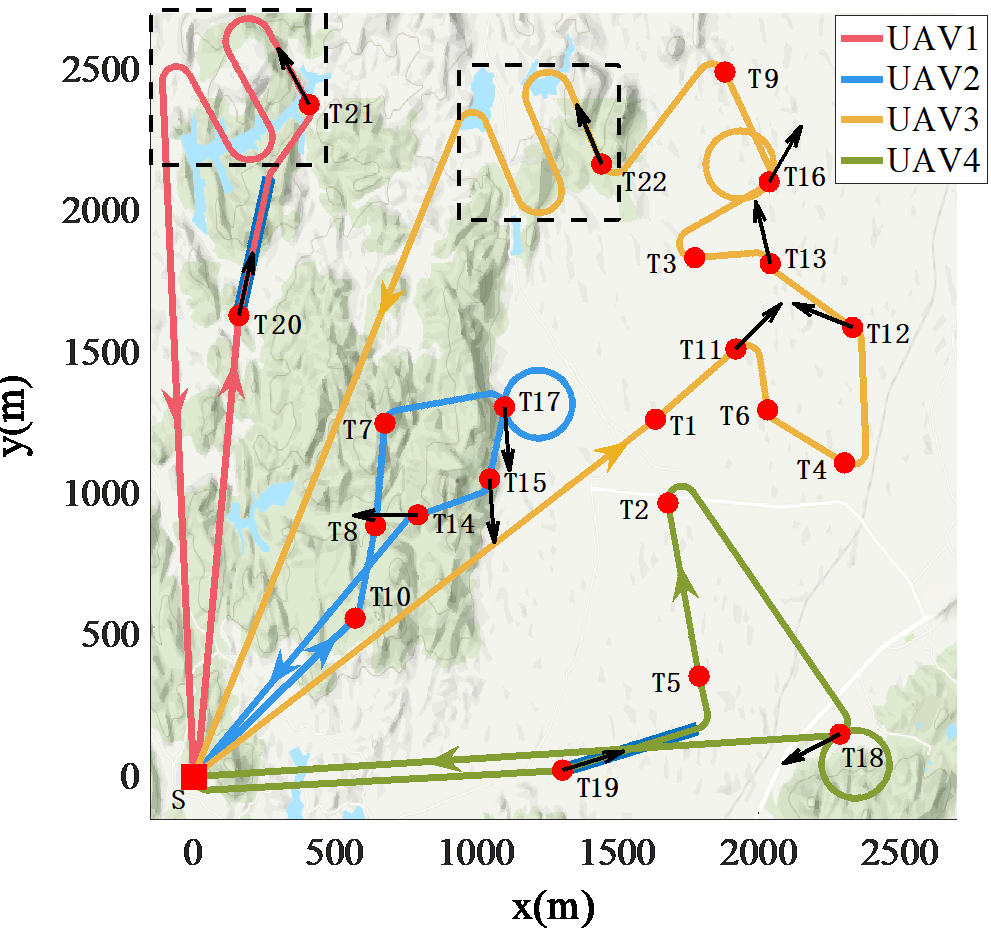}
\caption{Schematic diagram of typical application scenarios of the algorithm.}
\label{myfigure14}
\end{figure}

\begin{figure*}[h]
\centering
\includegraphics[width=1\linewidth]{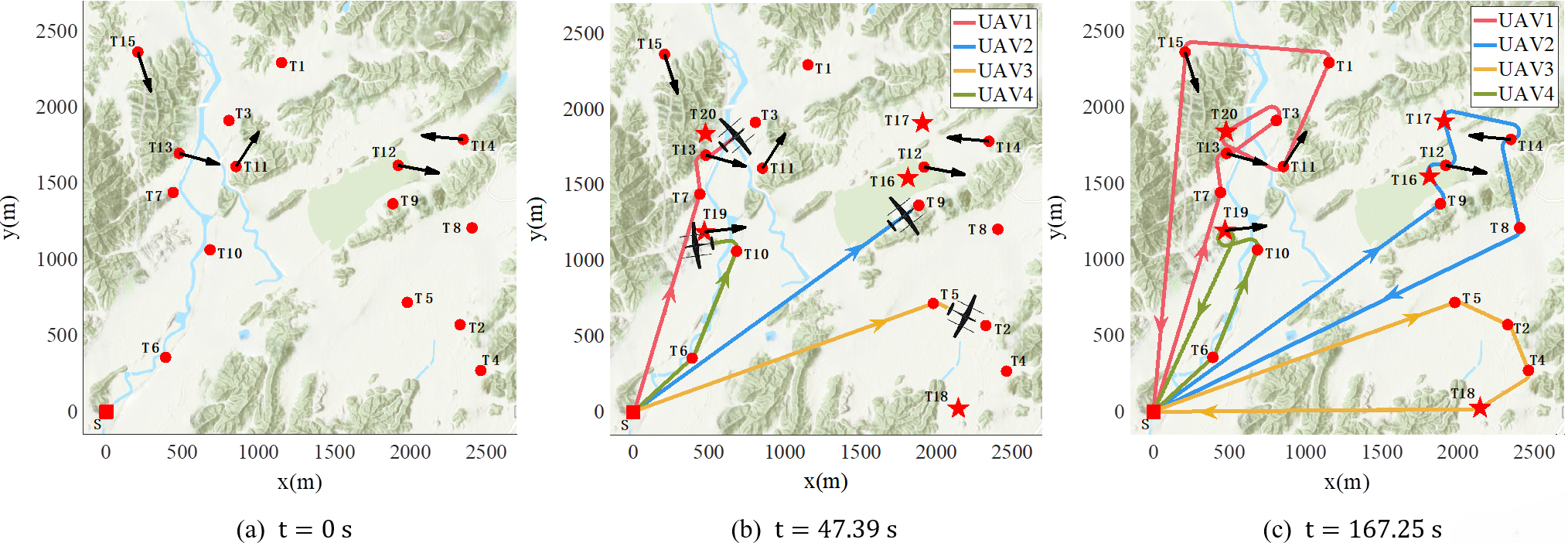}
\caption{Completion of tasks by UAV swarm at different times as new tasks emerge. Red pentagrams indicate emerging tasks.}
\label{myfigure15}
\end{figure*}

 Fig. \ref{myfigure14} provides a comprehensive illustration of a typical application of the proposed algorithm. First, it is apparent from the figure that the tasks performed by the four UAVs are clearly grouped into clusters. Different UAVs execute distinct task clusters. The preprocessing of task clustering introduces differentiation in task allocation, leading to significant variations in the number of tasks performed and the distance traveled by UAVs. 
 Fig. \ref{myfigure14} illustrates the ability of the UAVs to perform five heterogeneous tasks. Tasks 16-18 are circular tasks, tasks 19 and 20 are linear tasks, and tasks 21 and 22 are area-based tasks. These heterogeneous tasks are effectively performed by different UAVs. This is attributed to the adopted CS-type and CSC-type Dubins paths. Both path types fully consider task outgoing angle constraints, ensuring that the UAV’s path adheres to kinematic constraints without abrupt changes in direction.
 CSC-type paths also account for tasks with incoming angle constraints, such as straight-line targets, area targets, and others. This allows the UAV to smoothly comply with the heading angle constraints upon reaching the target position, enabling it to immediately begin executing the mission without the need for transition paths. 
 Notably, linear and area missions cause significant changes in the UAV’s position, moving it away from the original mission location. This poses a considerable challenge for static planning. However, our proposed algorithm shows strong real-time performance. As a result, it employs the current UAV position and attitude to determine the next task and plan the path in real time.
 
\subsection{Uncertainty Simulations}
\subsubsection{New task}
\begin{figure}[h]
\centering
\includegraphics[width=0.85\linewidth]{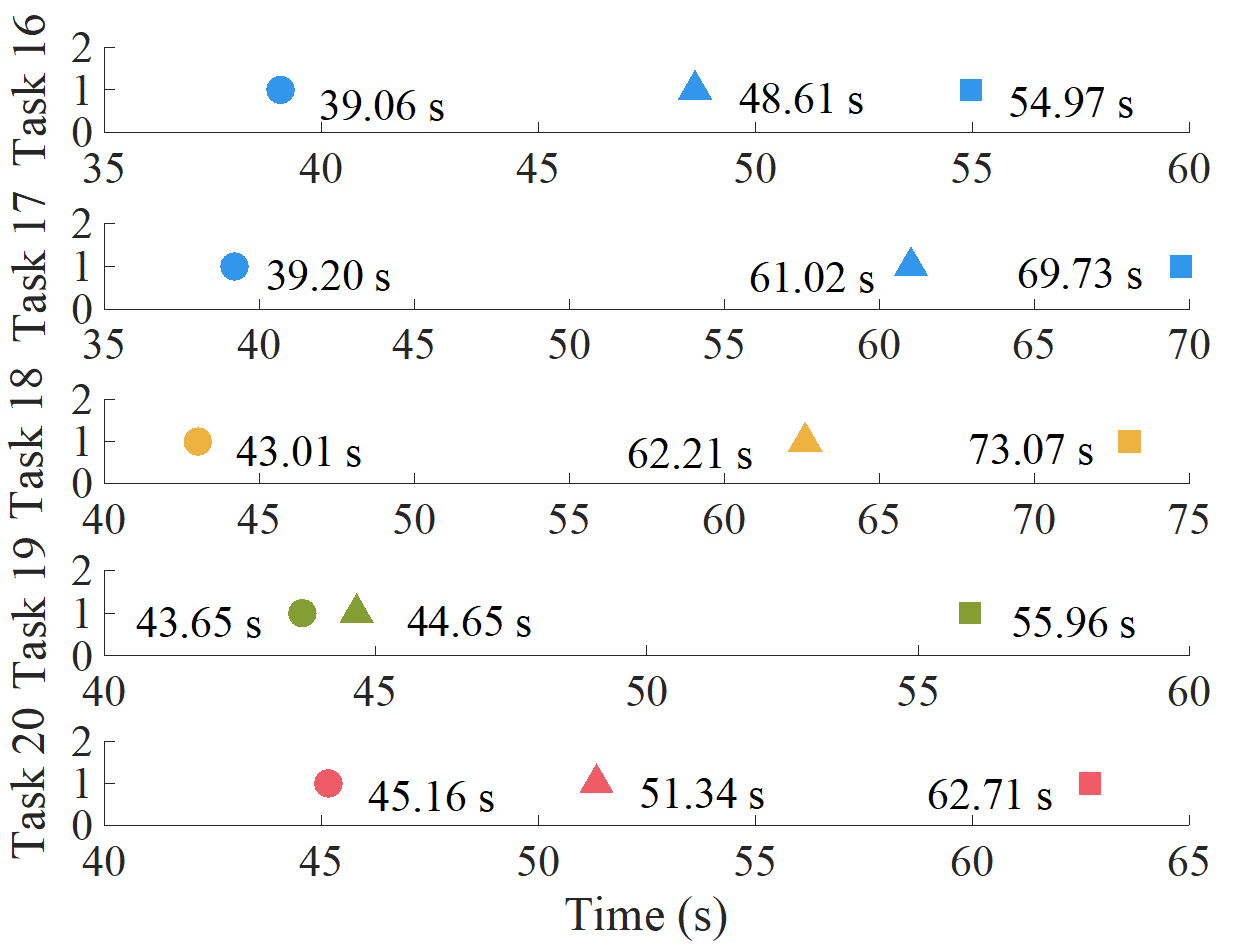}
\caption{New task appearance, allocation, and completion time chart. Task generation, assignment and completion are represented by circles, triangles and rectangles respectively. Red, blue, yellow and green indicate that the task was performed by UAVs 1, 2, 3 and 4 respectively}
\label{myfigure17}
\end{figure}

This section verifies that the proposed method can be applied to scenarios with new tasks. The base scenario in this section is the same as that in the previous section. The scenario involves four UAVs to accomplish 15 tasks randomly distributed in a 2.5km*2.5km area. The missions consist of 10 unconstrained point targets and 5 constrained point targets, which can be accomplished by any of the UAVs, as shown in Fig. \ref{myfigure15}a.
Circle, line and area tasks are similar to constrained point targets, with the addition of coverage paths. 
Therefore, these tasks apply point targets with heading angle constraints instead to simplify the problem.
New tasks are randomly generated within the range of motion during the time period $[30s, 50s]$.
Similarly, the new tasks are restricted to the unconstrained point targets and constrained point targets. It is assumed that none of the new tasks are urgent. 
Fig. \ref{myfigure15} displays the UAV swarm performs the mission at different moments in a scenario where new tasks emerge.
There are five new emerged tasks, four of which are general point tasks and one constrained point targets. The new tasks are assigned to the corresponding UAVs based on the clustering criterion when they appear. Eventually, the total number of tasks increases to 20. Fig. \ref{myfigure17} not only presents the generation, assignment, and completion times for each new task, but also indicates which UAV executes each task. From the time plot, it can be seen that the assignment of non-urgent new tasks has a confidential relationship with both the task generation time and the task generation location.


\subsubsection{UAV Damage}
\begin{figure}[ht]
\centering
\subfigure[]{
\includegraphics[width=0.7\linewidth]{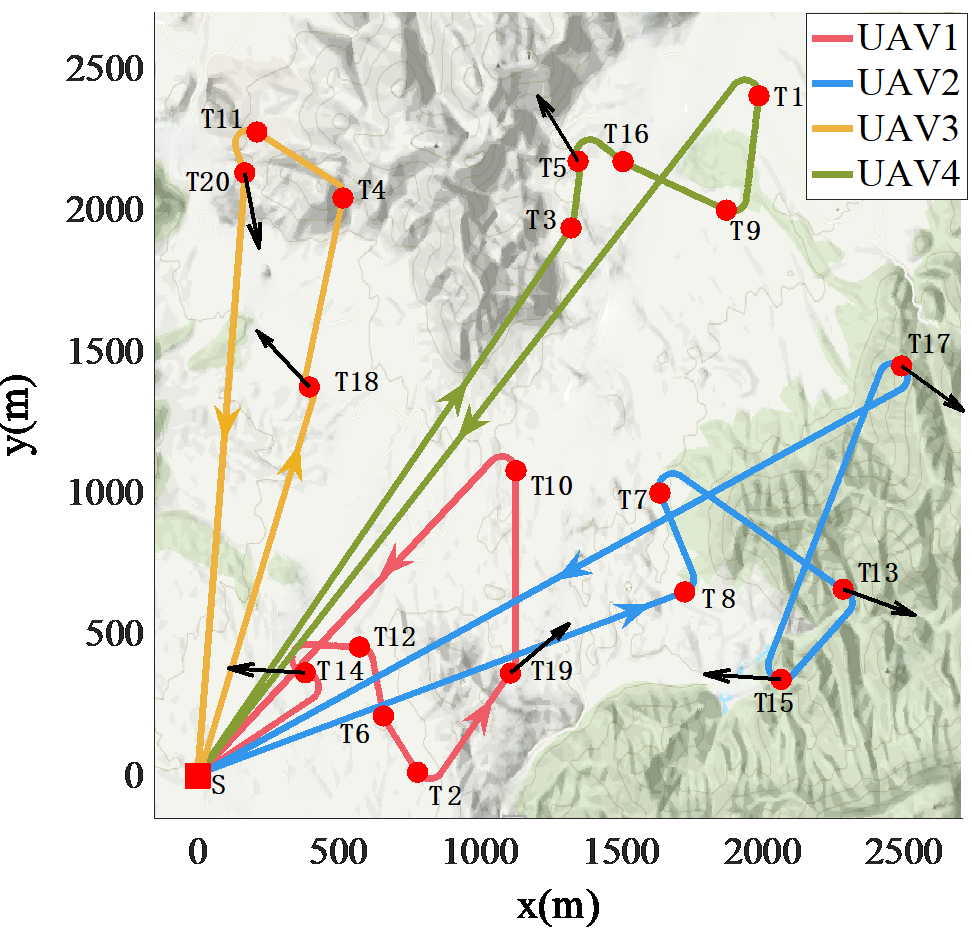}
\label{myfigure18}
}
\subfigure[]{
\includegraphics[width=0.7\linewidth]{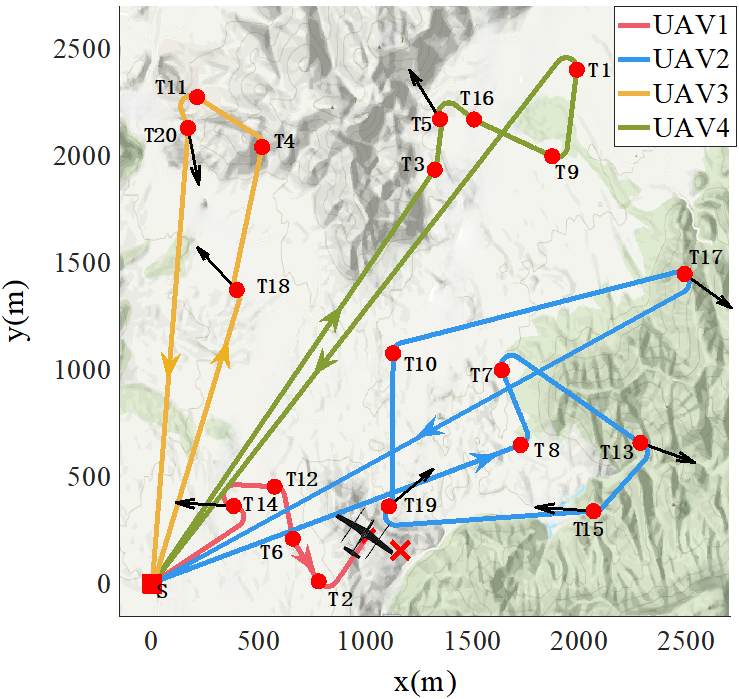}
\label{myfigure19}
}

\caption{(a) Planning diagram of the missions without UAV damage. (b) Planning diagram of missions with UAV damage. The red cross next to the UAV indicates that the UAV is damaged.}
\end{figure}

This section verifies that the proposed method can be adapted to UAV damage scenarios. The scenario involves four identical UAVs tasked with completing 20 tasks randomly distributed within a 2.5 km × 2.5 km area. The tasks consist of 12 unconstrained point targets and 8 constrained point targets, all of which can be completed by any of the UAVs. Fig. \ref{myfigure18} illustrates the planning diagram of the tasks without UAV damage. From the figure, we can see the task distribution and the respective order of completion for each UAV distance. 
At the 50th secon, UAV 1 is destroyed and loses at coordinates (1079.91, 326.88).
Fig. \ref{myfigure19} displays the mission planning diagram under the UAV destruction. After UAV destruction, its remaining unfinished tasks, tasks 10 and 19, are reassigned to UAV 2 and inserted into the original task sequence, disrupting the order of task execution. Therefore, UAV 2 actually completed 7 missions. 
UAV 4 has 5 original tasks (1, 3, 5, 9, 16; sequence: 3-16-5-9-1), of which 1 (task 16) is a constrained task, and 4 were unconstrained. 
UAV 3 has 4 original tasks (4, 11, 18, 20; sequence: 18-4-20-11), of which 2 (tasks 18 and 20) are constrained and 2 are unconstrained. 
Similarly, the reassignment of the remaining tasks from the destroyed UAV is closely related to both the time of its destruction and the locations of the tasks. 

\begin{table}[htbp]
    \begin{center}
    \caption{Comparison of tasking by UAV with and without UAV damage.}
    \label{mytable10}
    \begin{tabular}{lcc}
        \toprule
         UAV number  & UAV normal & UAV damage\\
        \midrule
        1 & \makecell[c]{5: T3→T16\\→T5→T9\\→T1} & \makecell[c]{5: T3→T16\\→T5→T9\\→T1}  \\ 
        2 &  \makecell[c]{6: T14→T12\\→T6→T2\\ \textcolor{red}{→T19→T10}} & \makecell[c]{4: T14→T12\\→T6→T2}  \\ 
        3 & \makecell[c]{4: T18→T4\\ →T20→T11} & \makecell[c]{4: T18→T4\\ →T20→T11}  \\ 
        4 & \makecell[c]{5: T8→T7 \\ →T13→T15 \\ →T17} &  \makecell[c]{5: T8→T7 \\ →T13→T15 \\ \textcolor{red}{→T19→T10} \\ →T17}  \\ 
        \bottomrule
    \end{tabular}
    \end{center}
\end{table}

\subsubsection{Extreme Scenario Simulation}
\begin{figure*}[h]
\centering
\includegraphics[width=1\linewidth]{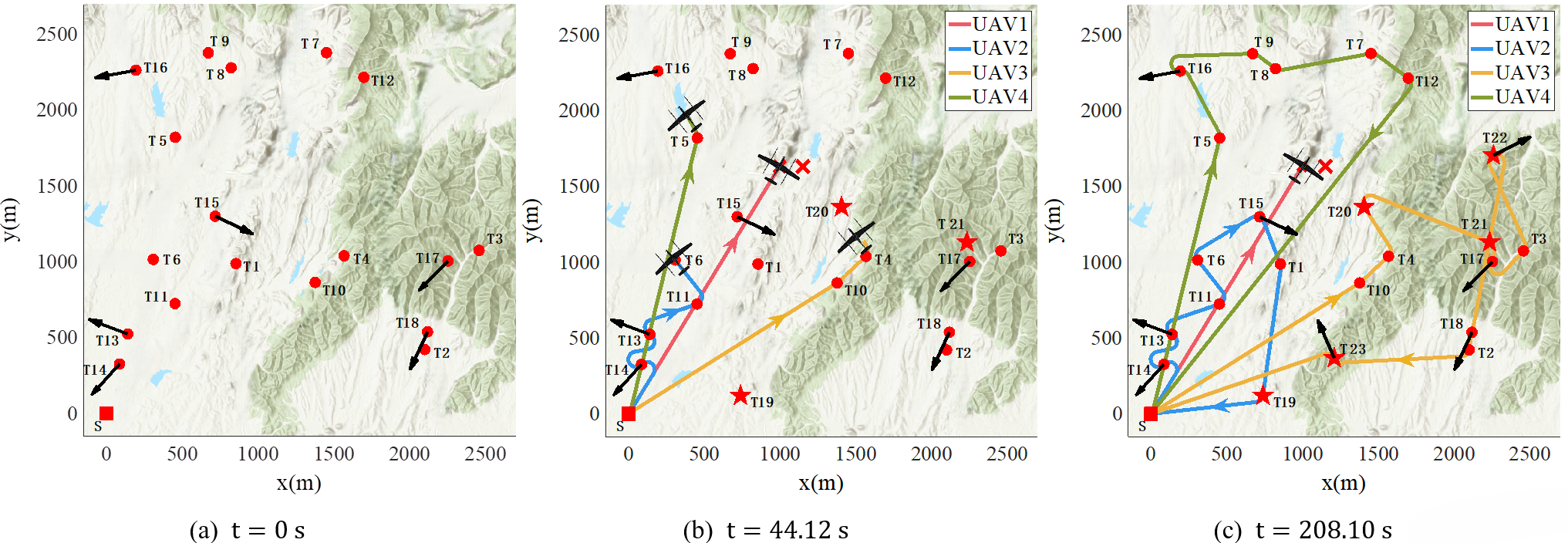}
\caption{Completion of tasks by UAV swarm at different times when two emergency situations occur simultaneously. Red pentagrams indicate emerging tasks. The red cross next to the UAV indicates that the UAV is damaged.}
\label{myfigure20}
\end{figure*}

\begin{figure}[h]
\centering
\includegraphics[width=0.85\linewidth]{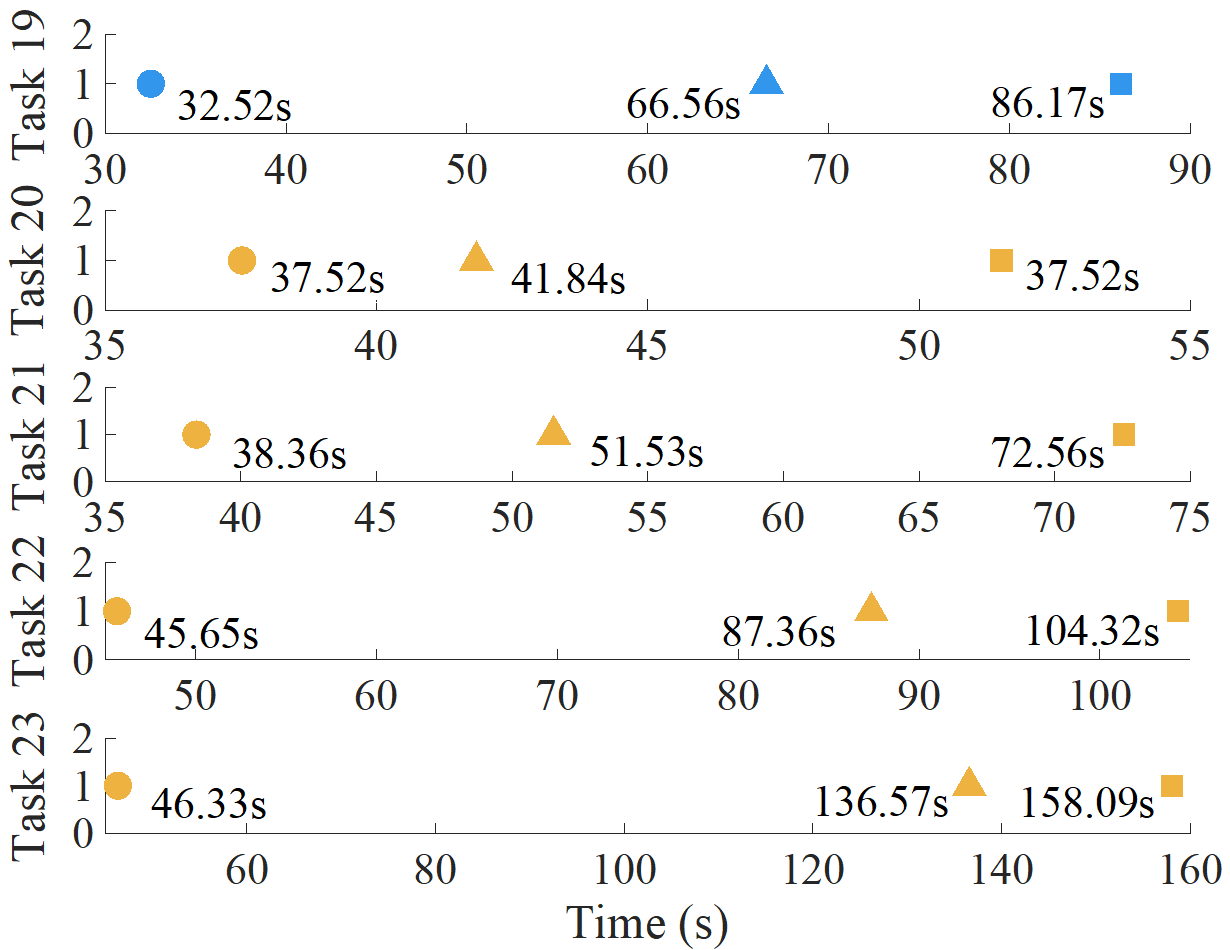}
\caption{New Task Appearance, Allocation, and Completion Time in the Extreme Scenario. Task generation, assignment and completion are represented by circles, triangles and rectangles respectively. Blue, yellow indicate that the task was performed by UAVs 2 and 3, respectively}
\label{myfigure22}
\end{figure}


This section demonstrates that the proposed method can be applied to a composite scenario where new tasks emerge and UAV destruction occurs simultaneously. 
The scenario involves 4 UAVs completing 18 tasks randomly distributed within a 2.5 km × 2.5 km area. The tasks consist of 12 unconstrained point targets and 6 constrained point targets, as shown in Fig. \ref{myfigure20}a.
To simplify the problem, new tasks are randomly generated within the range of motion at time period $[30s.50s]$.
Similarly, the new tasks are limited to unconstrained point targets and constrained point targets. It is assumed that none of the new tasks are urgent and one of the four UAVs is destroyed at a random time. 
Figs. \ref{myfigure20}a-c present the UAV swarm performs its mission at different times when both contingencies exist simultaneously.
UAV 1 is destroyed at (11510,60,989.19) in the 40th second.
Before the destruction of the UAV, a total of three unconstrained point targets appear. These are assigned to UAV 2 and 3.
UAV 1 is en route to task 7 when it is destroyed. Afterward, tasks 7 and 12, to which it was originally assigned, is reassigned to UAV 4.
After the UAV is destroyed, two constraint point targets appeared, all of which are assigned to UAV 3.
Fig. \ref{myfigure22} shows the generation, assignment, and completion times for each new task in the extreme scenario, and also indicates the UAVs that perform each task.

\section{Conclusions and Future Work}
This paper proposes a dynamic, real-time, multi-UAV cooperative mission algorithm, which is slightly improved to handle emergency situations.
The algorithm has four distinctive features: distributed mission planning, preprocessing of task clustering by proximity criterion, adoption of Dubins path distance cost and an alternative low-complexity task assignment strategy. 
These technical features allow the algorithm to perform real-time, coupled task assignment and path planning 
under various constraints such as heterogeneous tasks, nonholonomic kinematic constraints, time constraints, emergency, while making the total distance as short as possible. 
The following conclusions are drawn from the experiments. 

1)The distance cost function comparison experiments demonstrate that the time magnitude of the adopted Dubins path construction method is the same as that of the Euclidean path. Furthermore, this confirms that reducing the types of Dubins paths and simplifying their composition can significantly enhance computational efficiency. 

2)Assignment comparison experiment illustrates that mission planning algorithms applying different distance costs result in different allocation schemes. 
Further, experiments reveals the mechanism by which the additional distance caused by heading angle correction affects task allocation, as well as the conditions under which it is effective.

3)Spatial comparison experiments not only demonstrate that optimal task planning involves differentiated task allocation, but also verify that the proposed method can obtain suboptimal solutions.
The time comparison experiments suggest that both the total planning time and the individual planning time of the proposed algorithm are the shortest among the 7 algorithms, achieving mission planning on the order of $10^{-4}$ second, with a small sacrifice in path length.

4)Three distinct emergency experiments validate the effectiveness of key techniques, such as real-time mission planning of new tasks, and mission replanning when the number of available UAVs decreases.
Furthermore, the experiments demonstrate the excellent robustness and adaptability of the variants based on the proposed algorithm. 

Even though the proposed algorithm provides encouraging preliminary results, it relies on certain simplifying assumptions that warrant further investigation, thereby outlining key directions for future works.
Future work will focus on resolving potential conflicts caused by obstacles, which will make our planning safer. And the constraints of UAV range will also be considered.
To better validate the developed algorithm, another important aspect to consider is evaluating the efficiency of the algorithm in situations with a large number of tasks and highly uneven distribution.

\end{document}